\documentclass[lettersize,journal]{IEEEtran}
\usepackage{amsmath,amsfonts}
\usepackage{algorithmic}
\usepackage{array}
\usepackage[caption=false,font=normalsize,labelfont=sf,textfont=sf]{subfig}
\usepackage{textcomp}
\usepackage{stfloats}
\usepackage{url}
\usepackage{verbatim}
\usepackage{graphicx}
\usepackage{booktabs}
\usepackage{graphicx}
\usepackage{pifont}
\usepackage{amssymb}  
\usepackage{booktabs}
\usepackage{algorithm}
\usepackage{algorithmic}
\usepackage{multirow}
\usepackage{xcolor}  

\def\BibTeX{{\rm B\kern-.05em{\sc i\kern-.025em b}\kern-.08em
    T\kern-.1667em\lower.7ex\hbox{E}\kern-.125emX}}
\usepackage{balance}

\newcommand{\hidden}[1]{{}}
\newcommand{\std}[1]{\scriptsize{$\pm{}${#1}}}
\newcommand{\cmark}{\textcolor{green!60!black}{\checkmark}}
\newcommand{\xmark}{\textcolor{red}{\ding{55}}}  

\newcommand{\rev}[1]{{#1}}

\begin{document}
\title{Human-Aligned Procedural Level Generation Reinforcement Learning\\via Text-Level-Sketch Shared Representation}
\author{
In-Chang Baek\IEEEauthorrefmark{1},
Seoyoung Lee\IEEEauthorrefmark{1},
Sung-Hyun Kim,
Geumhwan Hwang,
Kyung-Joong Kim\IEEEauthorrefmark{2}
\\
Gwangju Institute of Science and Technology (GIST), Gwangju, South Korea \\
\texttt{\{inchang.baek,seoyoung.john\}}@gm.gist.ac.kr, kjkim@gist.ac.kr
\thanks{\IEEEauthorrefmark{1}Equal contribution \IEEEauthorrefmark{2}Corresponding author}

}

\maketitle

\begin{abstract}
Human-aligned AI is a critical component of co-creativity, as it enables models to accurately interpret human intent and generate controllable outputs that align with design goals in collaborative content creation.
This direction is especially relevant in procedural content generation via reinforcement learning (PCGRL), which is intended to serve as a tool for human designers. However, existing systems often fall short of exhibiting human-centered behavior, limiting the practical utility of AI-driven generation tools in real-world design workflows.
In this paper, we propose \textbf{VIPCGRL} (\textbf{V}ision-\textbf{I}nstruction \textbf{PCGRL}), a novel deep reinforcement learning framework that incorporates three modalities—text, level, and sketches—to extend control modalities and enhance human-likeness. We introduce a shared embedding space trained via quadruple contrastive learning across modalities and human-AI styles, and align the policy using an auxiliary reward based on embedding similarity.
Experimental results show that VIPCGRL \rev{achieves improved performance compared to} existing baselines in human-likeness, as validated by both quantitative metrics and human evaluations.
The code and dataset are available at https://github.com/bic4907/VIPCGRL.
\end{abstract}

\begin{IEEEkeywords}
procedural content generation, reinforcement learning, shared representation, multi-modal learning, human alignment
\end{IEEEkeywords}

\section{Introduction}

\IEEEPARstart{G}{ame} level design, a key aspect of game content, has been actively studied in the machine learning literature as part of efforts to automate game development.
The procedural content generation via machine learning (PCGML) literature \cite{summerville2018pcgml} has explored a range of generative approaches, including search-based methods, evolutionary algorithms, and deep reinforcement learning (DRL).
These generative models have been formulated in various ways, from offline end-to-end generation to mixed-initiative content creation frameworks \cite{delarosa2021mixed}.
In particular, in co-creativity settings where humans--AI collaborate, it is crucial that each party understands the partner's intentions and acts in alignment with humans.
\rev{While controllability enables users to specify desired properties, it does not inherently guarantee that the generated levels resemble those created by human designers.
As a result, achieving both controllability and human-likeness remains a central challenge in procedural content generation.}

Procedural content generation via RL (PCGRL, \cite{khalifa2020pcgrl}) is a data-free methodology for generating game content.
Recent studies have extended the controllability of PCGRL through various means, including vector-based condition inputs \cite{earle2021learning}, applications to 3D domains \cite{jiang2022learning}, game balancing mechanisms \cite{jeon2023raidenv}, scalability and generalization \cite{earle2024scaling}, reward generation \cite{baek2024chatpcg,baek2025pcgrllm}, and text-conditioned generation \cite{baek2025ipcgrl}.
However, current approaches remain limited to numerical condition values and textual inputs, lacking consideration of control modalities in content creativity. 
\rev{Additionally, even when generated levels satisfy specified conditions, they may still fail to capture the structural and stylistic properties of human-authored content, limiting their effectiveness in co-creative scenarios.}

\begin{figure}[]
    \includegraphics[width=1.0\linewidth]{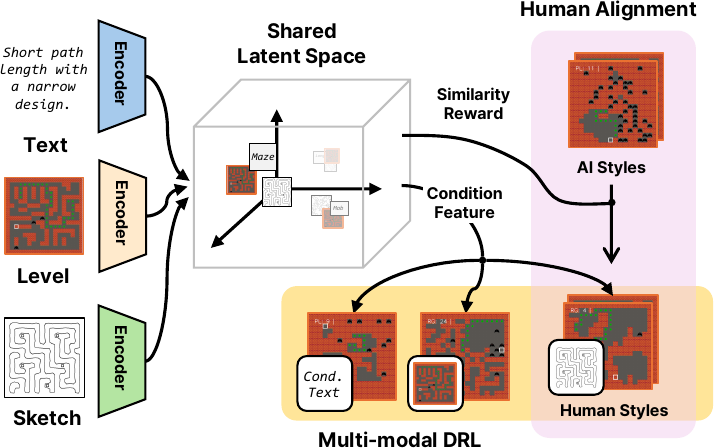}
    \caption{\textbf{Overview of VIPCGRL.} The text-level-sketch shared latent space enables multi-modal conditional DRL and aligns the policy with human styles.}
    \label{fig:teasure}
\end{figure}

\rev{This limitation highlights a fundamental gap in current PCGRL frameworks: while they are effective at optimizing task-specific objectives, they do not explicitly incorporate mechanisms for aligning generated content with human design styles.
In particular, existing approaches lack a unified representation that captures both multi-modal control signals and human-specific characteristics.
Addressing this gap requires a framework that can simultaneously (1) support diverse input modalities for intuitive control, and (2) align the generation process with human-authored examples.}

\rev{To address these challenges, we propose \textbf{VIPCGRL} (\textbf{V}ision-\textbf{I}nstruction PCGRL, Fig. \ref{fig:teasure}), a multi-modal DRL framework for human-aligned level generation.
Our approach integrates textual descriptions, level representations, and sketch-based inputs into a shared latent space, enabling expressive multi-modal control.
In addition, we introduce a similarity-based reward that encourages the agent to generate levels closer to human-designed examples in the learned representation space.
Through this design, VIPCGRL extends PCGRL from condition-driven generation to human-aligned generation.}

\rev{A key component of our framework is a shared latent representation learned through quadruple contrastive learning \cite{akbari2021vatt,radford2021learning}, which aligns multiple modalities while distinguishing between human- and AI-generated content.
This representation serves as a unified interface for both policy conditioning and reward computation, enabling consistent interpretation of text, level, and sketch inputs.
We further derive a similarity-based auxiliary reward from this space, encouraging the policy to generate levels closer to human-designed examples.
By tightly coupling representation learning with policy optimization, the agent learns to align its behavior with human design patterns.}

\rev{To systematically evaluate our approach, we focus on two key aspects: (1) alignment with human design styles, and (2) effectiveness of multi-modal conditional control.}
We evaluate the proposed method on a 2D level generation task, measuring performance in terms of controllability and human-likeness.
The quantitative results show that our method outperforms existing baselines in human-likeness and successfully incorporates text- and vision-based inputs.
Furthermore, human evaluations practically validate that our method achieves higher human-likeness and more coherent multi-modal conditional generation.

\rev{In summary, our contributions are as follows:
\begin{itemize}
\item We propose a DRL framework for PCGRL that explicitly optimizes human-likeness through a similarity-based alignment reward, enabling human-aligned level generation.
\item We introduce a shared latent representation that integrates text, level, and sketch modalities, allowing consistent multi-modal conditional control.
\item We introduce a quadruple contrastive learning objective that jointly aligns modalities and captures human–AI stylistic differences within a unified embedding space.
\item We demonstrate that our approach improves human-likeness while maintaining controllability, supported by both quantitative evaluation and human studies.
\end{itemize}}

\begin{figure}[]
    \centering
    \includegraphics[width=0.9\linewidth]{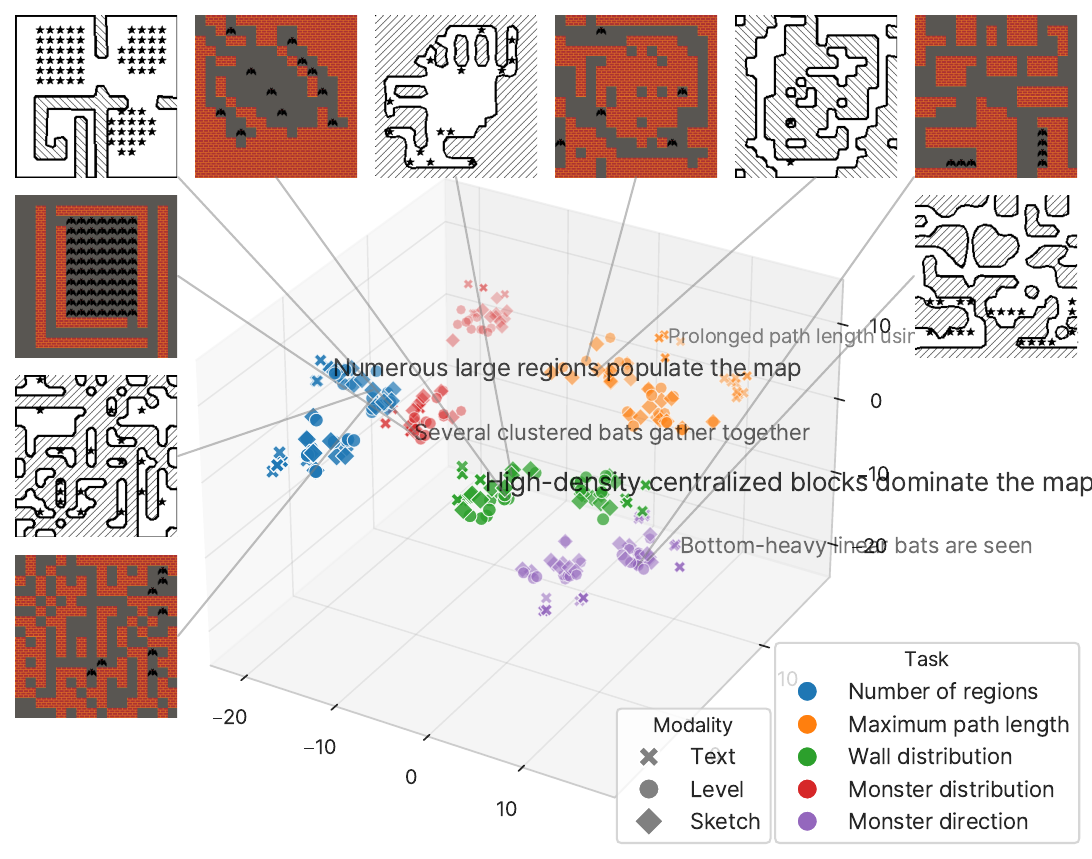}
    \caption{\textbf{Shared Latent Space Visualization.}
    The VIPCGRL encoder learns 64-dimensional representations for diverse instructions and modalities, which are projected into three dimensions using t-SNE \cite{maaten2008visualizing}.
    The different task embeddings are clustered within their respective inter-clusters, and the multi-modal embeddings corresponding to the same condition are also grouped within the same intra-cluster.
    }
    \label{fig:shared_rep}
    \vspace{-0.5cm}
\end{figure}

\section{Related Work}

\subsection{Procedural Game Level Generation via Reinforcement Learning}
\label{sec:bg_pcgrl}

PCGRL \cite{khalifa2020pcgrl} is a machine learning-based content generation method.
The benefits of PCGRL stem from its data-free nature and computational efficiency during inference, making it well suited for real-time content generation in games \cite{togelius2011search} and enabling co-creation with human designers \cite{delarosa2021mixed}.
PCGRL frames the level design process as a Markov Decision Process (MDP), where the game level is represented as a state \( s_t \), the level-revising operation as an action \( a_t \), and the agent receives a reward \( r_t \) according to how well it satisfies the given condition.

Controllable PCGRL (CPCGRL, \cite{earle2021learning}) employs a conditional vector and a parameterized reward function to train conditional DRL agents. \rev{This approach has been validated across various domains, including 2D environments \cite{earle2021learning}, 3D environments \cite{jiang2022learning}, and graph data generation \cite{rupp2024g}.} While effective, these methods rely solely on pre-defined scalar inputs.
Instructed PCGRL (IPCGRL, \cite{baek2025ipcgrl}) extends CPCGRL by incorporating a BERT-based text encoder, enabling natural language-conditioned generation. Multi-objective instructed PCGRL (MIPCGRL, \cite{kim2025multi}) further improves language-conditioned PCGRL by learning objective-aware representations for instructions containing multiple generation objectives.
However, two key limitations remain: (1) the text-only modality limits the expression of spatial intent; and (2) the reliance on environmental rewards alone in DRL training causes a stylistic \rev{mismatch with human-designed levels}, ultimately hindering co-creative alignment.
\rev{These limitations highlight the need for a framework that supports richer input modalities while explicitly incorporating human design characteristics into the generation process.}
The comparison of prior methods and our approach is provided in Table~\ref{tab:previous_methods}.

\begin{table}[]
\centering
\caption{Comparison between prior controllable PCGRL methods and VIPCGRL in terms of input modality support and human alignment. (\textit{T}: text, \textit{V}: vision)}
\label{tab:previous_methods}
\resizebox{\columnwidth}{!}{%
\begin{tabular}{lcccc}
\toprule
 & Instruction (\textit{T}) & Level (\textit{V}) & Sketch (\textit{V}) & Human-Aligned \\ \midrule
CPCGRL \cite{earle2021learning} & \xmark & \xmark & \xmark & \xmark \\
IPCGRL \cite{baek2025ipcgrl} & \cmark & \xmark & \xmark & \xmark \\
MIPCGRL \cite{kim2025multi} & \cmark & \xmark & \xmark & \xmark \\
\textbf{VIPCGRL} & \cmark & \cmark & \cmark & \cmark \\ 
\bottomrule
\end{tabular}
}
\end{table}

\textbf{Environment.}
The 2D level generation task is a widely used benchmark environment for evaluating DRL-based generative policies \cite{khalifa2020pcgrl,earle2021learning,earle2024scaling}.
The environment is represented as an $N^{2}$ matrix, where each cell corresponds to a game tile (e.g., empty, wall, monster) or the agent’s position.
The action space consists of seven discrete actions that allow the agent to traverse the level and modify tile types.
The environment includes five independent tasks, each defined by a specific structure or semantic constraint: number of regions, maximum traversable path length, wall distribution, the spatial distribution of monster tiles, and the directional distribution of tiles.
The agent is trained using environment-based rewards that reflect the difference between the generated level and the target conditions, receiving positive or negative feedback based on task-specific loss.
The details of the environment and the tasks are described in Appendix~VIII.

\begin{figure*}[!]
    \centering
    \includegraphics[width=1.0\linewidth]{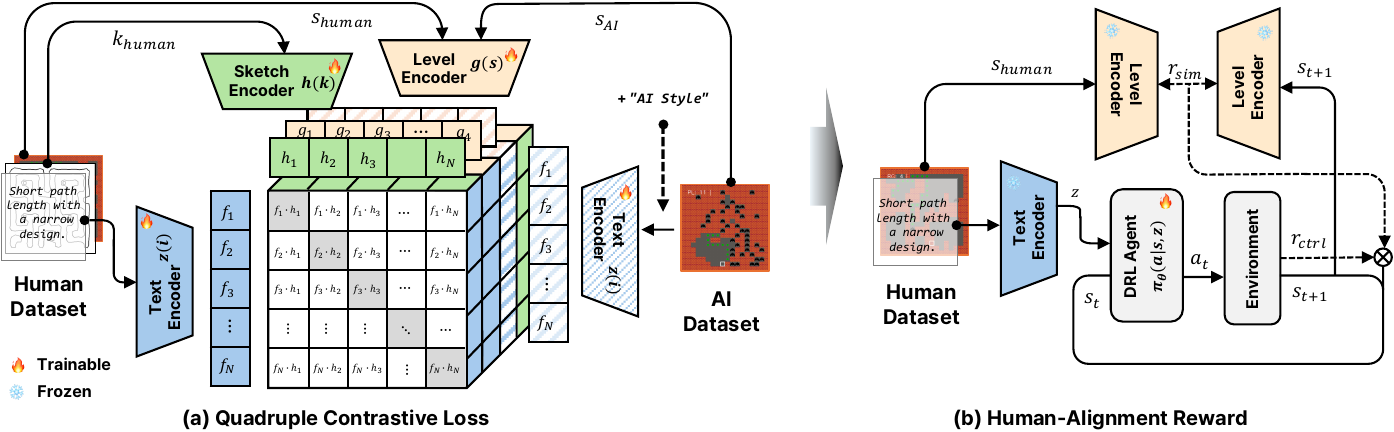}
    \caption{\textbf{VIPCGRL Training Procedure.} The proposed method follows a two-step training process: (a) learning a shared representation via multi-modal human-AI contrastive learning, and (b) training a controllable DRL agent using conditional input and a similarity-based reward. The dashed background indicates contrastive learning performed on the AI-generated level dataset.}
    \label{fig:method}
    \vspace{-0.4cm}
\end{figure*}

\subsection{Shared Embedding Space Representation via Contrastive Learning}
\label{sec:bg_shared_latent}
\rev{Contrastive learning has been widely used to learn shared representations across multiple modalities by aligning semantically related data in a common embedding space.}
Foundational work such as CLIP \cite{radford2021learning} aligned images and text through contrastive learning on a vast dataset, demonstrated strong zero-shot classification capabilities.
\rev{Subsequent approaches have extended this paradigm to support multiple modalities, enabling unified representations across diverse data types \cite{guzhov2022audioclip, wang2023connecting, girdhar2023imagebind}.
While these methods successfully capture semantic correspondence across modalities, they do not explicitly model stylistic differences between data sources, such as human- and AI-generated content.}

This semantic grounding is also being increasingly leveraged in reinforcement learning. 
\rev{Recent work uses contrastive representations as reward signals by measuring similarity between goal descriptions and current observations \cite{fu2024,rocamonde2024vision,du2023vision}, with applications in navigation and robotic manipulation \cite{majumdar2022zson, nair2022r3m}.}
However, these methods typically rely on pre-trained, frozen encoders, a limitation that confines their focus to task completion. 
\rev{These limitations highlight the need for a representation learning framework that not only aligns multiple modalities but also captures human-specific stylistic characteristics within the embedding space.}

\section{Methodology}
\label{sec:method}
Key challenges in multi-modal PCGRL include achieving the integration of multiple modalities while preserving controllability, and learning human-like suboptimal policies that can satisfy task-specific conditions.
The proposed VIPCGRL framework is composed of two training steps, as illustrated in Fig.~\ref{fig:method}. 
The shared representation, illustrated in Fig. \ref{fig:shared_rep}, is trained using a contrastive learning method, and the pre-trained latent space is utilized for multi-modal conditional DRL and for learning a human-aligned policy.

\subsection{Dataset}
We construct a dataset consisting of text-level-sketch triples to train encoder models that project the three modalities into a shared latent space.
Each sample is represented as a triple $(i, s, k)$, where $i$ denotes a natural language instruction, $s$ is a level state, and $k$ is a sketch image that created from the level image.
To enable multi-modal capability and align text with level-sketch representations, we collect a human-designed 2D level dataset $\mathcal{D}_\mathrm{human}$ of 5K samples in the environment \cite{earle2021learning} from four game level designers, where each sample consists of a text instruction and a level pair.
The human instruction set consists of 160 instructions distributed across five level design tasks, with 40 structural conditions—pertaining to quantity, shape, and distribution—assigned to each task.

The human-designed level states $S^\mathrm{human}$, each with a shape $(16, 16, 3)$, are paired with the given instructions, where 16 denotes the width and height, and 3 represents the one-hot encoded tile types.
These level states are then transformed into sketch-style images using a stylization function $f_{\text{style}}$, which produces a conceptual visual abstraction of the level.
The stylization function applies standard computer vision techniques, including edge detection, spline interpolation, and Gaussian blurring. The resulting grayscale sketches have a resolution of $(224, 224, 1)$. 
Inspired by CoordConv \cite{liu2018intriguing}, we append two channels containing x-y coordinate information to both visual inputs (levels and sketches) to provide them with explicit spatial context.

To expose the model to diverse design styles beyond those in our human-designed data, we construct an AI-generated dataset $D_{AI}$, consisting of 2.5K AI-generated samples using a pretrained IPCGRL model~\cite{baek2025ipcgrl} with the instruction set $\mathcal{I}$ utilized in IPCGRL.
We then pair these generated levels with a new instruction set $\mathcal{I}^{AI}$, which was created by modifying 80 of the original instructions with prefixes such as “AI-style”. 
The primary stylistic distinction in our dataset is between human-designed and AI-generated levels, as they exhibit distinct design patterns.
The AI-style dataset is keyword-augmented to align with the human dataset's instructions.
Examples of the instruction sets are provided in Appendix~XIV.
The two datasets are formally defined as follows:
\begin{equation}
    \mathcal{D}_\mathrm{human} = (i \in \mathcal{I},\ s \in S^\mathrm{human}(i), \ k \in f_{\text{style}}({S^\mathrm{human}}(i)))
\end{equation}
\begin{equation}
    \mathcal{D}_{AI} = (i^{AI} \in \mathcal{I}^{AI},\ s \in S^{AI}(i))
\end{equation}

\begin{table*}[!t]
\small
\caption{\textbf{Generalization Performance Across Modalities.} This table presents VIPCGRL's inference-time performance across input modalities, with different similarity reward coefficient. \rev{Bold values indicate the best performance per metric.} Arrows indicate whether higher ($\uparrow$) or lower ($\downarrow$) values are preferred. (\textit{T}: text, \textit{V}: vision, *: unseen modality)}

\label{tab:quantitative_main}
\begin{tabular}{p{4.0cm}p{2.0cm}p{2.2cm}p{2.2cm}p{2.2cm}p{2.2cm}}
\toprule
 &  & Controllability & \multicolumn{2}{c}{Human-likeness} & Diversity \\
 & Input & Progress$\uparrow$ & TPKL-Div$\downarrow$ & ViT-Sim$\uparrow$ & Hamming$\uparrow$ \\
\midrule
Random & - & 0.02 \std{0.01} & 8.72 \std{0.08} & 0.73 \std{0.00} & \textbf{0.59} \std{0.01} \\
CPCGRL \cite{earle2021learning} & Text (\textit{T}) & \textbf{0.90} \std{0.03} & 7.57 \std{0.30} & 0.74 \std{0.00} & 0.46 \std{0.01} \\
IPCGRL \cite{baek2025ipcgrl} & Text (\textit{T}) & 0.87 \std{0.05} & 7.67 \std{0.41} & 0.74 \std{0.01} & 0.48 \std{0.01} \\
\midrule
 & Text (\textit{T}) & 0.88 \std{0.02} & 6.18 \std{0.32} & 0.75 \std{0.00} & 0.39 \std{0.01} \\
VIPCGRL \scriptsize{($\lambda_{\mathrm{sim}}=0.5$)} & Level* (\textit{V}) & 0.70 \std{0.01} & 6.17 \std{0.27} & 0.75 \std{0.00} & 0.40 \std{0.01} \\
 & Sketch* (\textit{V}) & 0.72  \std{0.02} & 6.24 \std{0.28} & 0.75 \std{0.00} & 0.40 \std{0.01} \\
\midrule
 & Text (\textit{T}) & 0.84 \std{0.05} & \textbf{5.53} \std{0.12} & \textbf{0.77} \std{0.00} & 0.41 \std{0.02} \\
VIPCGRL \scriptsize{($\lambda_{\mathrm{sim}}=1$)} & Level* (\textit{V}) & 0.67  \std{0.03} & 5.58 \std{0.10} & \textbf{0.77} \std{0.00} & 0.42 \std{0.02} \\
 & Sketch* (\textit{V}) & 0.70  \std{0.04} & 5.63 \std{0.13} & \textbf{0.77} \std{0.00} & 0.42 \std{0.02} \\
\bottomrule
\end{tabular}
\vspace{-0.4cm}
\end{table*}

\subsection{Learning Text-Level-Sketch Latent Space}
\label{sec:clip_latent}
To learn a unified latent representation from multi-modal inputs, we propose the quadruple contrastive loss. 
The term “Quadruple” reflects the four-fold alignment enforced by our method, which jointly aligns representations across three distinct modalities and a fourth key aspect: the creation style of human and AI-generated content.
We structure the contrastive learning process based on a multi-positive InfoNCE objective, where semantic criteria tailored for the PCGRL domain explicitly define the positive and negative pairs.
These criteria, defined during dataset construction, consist of structure conditions and human and AI-generated styles. 
Ultimately, the quadruple contrastive loss trains encoders to pull representations of multi-modal samples that sharing the same structure condition and style closer together in the latent space, while pushing apart those that differ in any of these aspects.

The latent embeddings of each modality are represented as 64-dimensional vectors, denoted by $z$, $g$, and $h$, respectively. The encoders are defined as follows:
\begin{equation}
z=f_\psi(i),\quad g=f_\phi(s),\quad h=f_\omega(k)
\end{equation}
Here, \(f_\psi\) uses a pre-trained CLIP text encoder with fixed weights, followed by a trainable single-layer MLP to project into the latent space, and \(f_\phi, f_\omega\) are separate trainable architectures each composed of a Residual CNN and a projection MLP. Denote the embedding of modality \(m\) as \(e^{(m)}\), with \(M=\{\text{text},\text{level},\text{sketch}\}\). The overall loss aggregates over all ordered modality pairs:
\begin{equation}
\mathcal{L}=\sum_{m\neq n}\mathcal{L}_{m\to n}
\end{equation}
\begin{equation}
\mathcal{L}_{m \to n}
=-\frac{1}{|B|} \sum_{i \in B}
\log
\frac{
    \sum\limits_{j \in P_{m \to n}(i)}
    \exp\left(
    \frac{
        \mathrm{sim}\left(
        e^{(m)}_i, \;
        e^{(n)}_j
        \right)
    }{\tau}
    \right)
    }{
    \sum\limits_{k \in B}
    \exp\left(
    \frac{
        \mathrm{sim}\left(
        e^{(m)}_i, \;
        e^{(n)}_k
        \right)
    }{\tau}
\right)}
\end{equation}
Here, $P_{m \to n}(i)$ denotes the set of positive samples in modality $n$ that share the same task and style as the sample $i$ in modality $m$. Each batch $B$ is constructed to include the same number of samples for all modalities, and human and AI style data are randomly sampled. All directional loss terms are summed with equal weights. The function $\mathrm{sim}(u,v)$ represents the cosine similarity, and $\tau$ is a learnable temperature parameter.
As shown in Fig. \ref{fig:shared_rep}, each encoder maps its respective modality into a semantically consistent space, where embeddings with the same condition are clustered together regardless of input type.
\rev{The Top-1 accuracy across modalities achieves an average of $0.98\pm{}0.01$, with relatively higher performance in text-level alignment compared to slightly lower accuracy in sketch-level alignment.}

\subsection{Conditional Reinforcement Learning with Human-Alignment Reward}
\label{sec:pcgrl_reward}
\paragraph{Multi-modal Condition} 
We train a conditional DRL agent using a policy $\pi(a \mid s, c)$, where separate models are trained for each environment task.
Here, $c \in \{z, h, g\}$ denotes a conditional feature from one of the three modalities: text ($z$), level ($g$), or sketch ($h$). To support this, we employ three frozen encoders: $f_{\psi}$, $f_{\omega}$, and $f_{\phi}$, each mapping its respective modality into a shared latent space. 
While only the text encoder $f_{\psi}$ is used for conditional feature extraction in this study, the framework allows training any one of the three encoders independently, offering flexibility with inference cost.

\newcommand{\imgrandomstepone}{\raisebox{-.5\height}{\includegraphics[width=1.35cm]{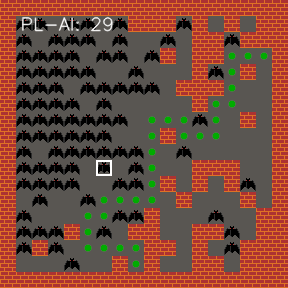}\includegraphics[width=1.35cm]{experiment/fig_prev_methods/media_images_Image_reward_99_seed_5_1989_3e64752640dd4cba29f4.png}}}

\newcommand{\imgrandomstepthree}{\raisebox{-.5\height}{\includegraphics[width=1.35cm]{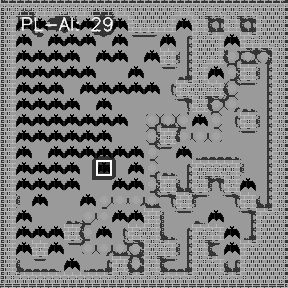}\includegraphics[width=1.35cm]{experiment/fig_prev_methods/AAF5y4fBR-w_1750352595284.png}}}

\newcommand{\oneimg}[1]{%
  \raisebox{-.5\height}{\includegraphics[width=1.35cm]{#1}}%
}

\newcommand{\twoimg}[2]{%
  \raisebox{-.5\height}{%
    \includegraphics[width=1.30cm]{#1}%
    \includegraphics[width=1.30cm]{#2}%
  }%
}

\newcommand{\rowA}{r16}
\newcommand{\rowB}{r22}
\newcommand{\rowC}{r48}
\newcommand{\rowD}{r53}
\newcommand{\textA}{Several small regi- ons are scattered}
\newcommand{\textB}{\mbox{Large regions show} \mbox{balanced distribution}}
\newcommand{\textC}{Moderate path length along a narrow path}
\newcommand{\textD}{Reasonable wide path offering balance}

\begin{table*}[!h]
{\small
\centering
\caption{\textbf{Multi-modal Generation Results.} The VIPCGRL agent is trained using text-conditioned inputs; however, the shared representation enables multi-modal generation during inference. Incorporating a higher similarity reward encourages the learning of policies that are more closely aligned with human design preferences.}
\label{tab:multi moal generation}

{
\renewcommand{\arraystretch}{-0.5}
\begin{tabular}{@{}p{2.57cm}p{2.57cm}p{2.57cm}p{2.57cm}p{2.57cm}p{2.57cm}@{}}
\toprule
Input & \multicolumn{2}{c}{VIPCGRL} & Input & \multicolumn{2}{c}{VIPCGRL} \\
 & $\lambda_{sim}=0.5$ & $\lambda_{sim}=1.0$ & & $\lambda_{sim}=0.5$ & $\lambda_{sim}=1.0$  \\ 

\midrule
\vspace{-0.0cm}\texttt{\tiny{\textA/ \textB}} & 
\vspace{-0.0cm}\twoimg{experiment/table2/\rowA/text_sim15.png}{experiment/table2/\rowB/text_sim15.png}  &
\vspace{-0.0cm}\twoimg{experiment/table2/\rowA/text_sim30.png}{experiment/table2/\rowB/text_sim30.png} &

\vspace{-0.0cm}\texttt{\tiny{\textC/ \textD}} & 
\vspace{-0.0cm}\twoimg{experiment/table2/\rowC/text_sim15.png}{experiment/table2/\rowD/text_sim15.png}  &
\vspace{-0.0cm}\twoimg{experiment/table2/\rowC/text_sim30.png}{experiment/table2/\rowD/text_sim30.png} \\

\vspace{0.25cm}\twoimg{experiment/table2/\rowA/state_input.png}{experiment/table2/\rowB/state_input.png}  &
\vspace{0.25cm}\twoimg{experiment/table2/\rowA/state_sim15.png}{experiment/table2/\rowB/state_sim15.png}  &
\vspace{0.25cm}\twoimg{experiment/table2/\rowA/state_sim30.png}{experiment/table2/\rowB/state_sim30.png} &

\vspace{0.25cm}\twoimg{experiment/table2/\rowC/state_input.png}{experiment/table2/\rowD/state_input.png}  &
\vspace{0.25cm}\twoimg{experiment/table2/\rowC/state_sim15.png}{experiment/table2/\rowD/state_sim15.png}  &
\vspace{0.25cm}\twoimg{experiment/table2/\rowC/state_sim30.png}{experiment/table2/\rowD/state_sim30.png} \\

\vspace{0.25cm}\twoimg{experiment/table2/\rowA/sketch_input.png}{experiment/table2/\rowB/sketch_input.png}  &
\vspace{0.25cm}\twoimg{experiment/table2/\rowA/sketch_sim15.png}{experiment/table2/\rowB/sketch_sim15.png}  &
\vspace{0.25cm}\twoimg{experiment/table2/\rowA/sketch_sim30.png}{experiment/table2/\rowB/sketch_sim30.png} &

\vspace{0.25cm}\twoimg{experiment/table2/\rowC/sketch_input.png}{experiment/table2/\rowD/sketch_input.png}  &
\vspace{0.25cm}\twoimg{experiment/table2/\rowC/sketch_sim15.png}{experiment/table2/\rowD/sketch_sim15.png}  &
\vspace{0.25cm}\twoimg{experiment/table2/\rowC/sketch_sim30.png}{experiment/table2/\rowD/sketch_sim30.png} \\ \bottomrule
 \\
\end{tabular}%
}
}

\end{table*}

The training process for a single task-specific agent is as follows. At the beginning of each episode, $i$ is randomly sampled from $\mathcal{I}_\mathrm{task}$, the subset of instructions corresponding to that specific task.
The instruction is embedded into a latent vector $z = f_{\psi}(i)$ using the frozen text encoder, which is then concatenated with the environment state and used as input to the policy. 
The projection of the three modalities into a shared latent space facilitates inference on previously unseen modalities without requiring further training.

\paragraph{Similarity Reward} 
Following prior work~\cite{earle2021learning,baek2025ipcgrl}, the control reward $r^{\mathrm{ctrl}}_t$ measures how well the current state $s_t$ satisfies the instruction-conditioned goal.
Let $v_{\mathrm{goal}}$ denote the target value (e.g., desired number of monsters), and $f_{\mathrm{metric}}(s)$ be a function computing the corresponding property from a state $s$. The reward is defined as the temporal improvement toward the goal:
\begin{equation}
r^{\mathrm{ctrl}}_t = \left|v_{\mathrm{goal}} - f_{\mathrm{metric}}(s_{t-1})\right|_{L1} - \left|v_{\mathrm{goal}} - f_{\mathrm{metric}}(s_t)\right|_{L1}
\end{equation}

Handcrafting stylistic differences through the reward function is a challenging problem.
To further align the agent's behavior with human design preferences, we introduce a human-similarity reward $R_{\mathrm{sim}}$. 
\rev{At each step, the agent samples a human-designed level state $s_{\mathrm{human}} \sim \mathcal{D}_{\mathrm{human}}$ in a condition-wise manner, conditioned on the same task and instruction as the current episode.}
The frozen level encoder $f_{\phi}$ embeds environment state $s_t$ and the human reference level $s_{\mathrm{human}}$ into a shared representation space.
\rev{The similarity reward is defined as the relative change in cosine similarity, capturing the semantic distance between the agent's level and a reference human-designed level:}
\begin{equation}
\small
r^{\mathrm{sim}}_t = \cos\left( f_{\phi}(s_t), f_{\phi}(s_{\mathrm{human}}) \right) - \cos\left( f_{\phi}(s_{t-1}), f_{\phi}(s_{\mathrm{human}}) \right)
\end{equation}

The conditional DRL agent is trained with the following reward function that encourages both task completion and human-likeness by combining the two components, where $\alpha=30$ is a scaling factor used to normalize the magnitude of the two reward components, and $\lambda_{\mathrm{sim}}$ is a coefficient parameter that balances control and alignment objectives:
\begin{equation}
r_t = r^{\mathrm{ctrl}}_t + \alpha \cdot\lambda_{\mathrm{sim}} \cdot r^{\mathrm{sim}}_t
\end{equation}

\section{Experiment}
\subsection{Hypothesis}
\rev{To evaluate the effectiveness of the proposed framework, we formulate the following research hypotheses (RHs):}
\begin{itemize}
    \item \rev{\textbf{Human Alignment (RH1):} The similarity-based auxiliary reward, derived from a shared latent space trained to distinguish human design styles, encourages generated levels to exhibit structural characteristics consistent with human-designed levels.}
    \item \rev{\textbf{Multimodal Control (RH2):} The shared latent representation enables simultaneous control over both textual and visual modalities while maintaining comparable controllability.}
\end{itemize}

\subsection{Setup}

\subsubsection{Metrics}
\label{sec:eval_metric}
\rev{To comprehensively evaluate controllability, human-likeness at both the coarse semantic and fine-grained local pattern levels, and diversity, we adopt a set of complementary quantitative metrics.}

\textbf{\textit{Progress}} \cite{earle2021learning} quantifies the controllability of the model, reflecting the extent to which the agent’s behavior aligns with the specified input condition.
It is measured as the relative improvement from the initial level state $s_0$ to the terminal state $s_T$ with respect to the goal condition $v_{\text{goal}}$.
Using this function, progress is computed as:
\[
\mathrm{Progress}(s_0, s_T, v_\mathrm{goal}) = 1 - \left| \frac{v_{\mathrm{goal}} - f_{\mathrm{metric}}(s_T)}{v_{\mathrm{goal}} - f_{\mathrm{metric}}(s_0)} \right|
\]

\textbf{\textit{TPKL-Div}} (Tile Pattern KL-Divergence, ~\cite{lucas2019tile}) measures the syntactic human-likeness by comparing local tile pattern distributions. Let $P(t)$ and $Q(t)$ denote the frequency of tile pattern $t \in \mathcal{T}$ in the human and generated levels, extracted using sliding windows of sizes 2$\times$2, 3$\times$3. The divergence is computed as:
\[
D_{\text{KL}}(P \parallel Q) = \sum_{t \in \mathcal{T}} P(t) \log \left( \frac{P(t)}{Q(t)} \right)
\]
We calculate both pattern-level and raw tile-level divergences and combine them with equal weight ($w = 0.5$). Lower values indicate higher alignment with human design in a statistical sense.

\textbf{\textit{ViT-Sim}} (Vision Transformer Similarity, ~\cite{zhu2024evaluate}) captures coarse semantic human-likeness comparing generated and human-designed levels. We render each level into an image and extract embeddings using a pre-trained ViT\footnote{We use the official pre-trained google/vit-base-patch16-224 model.}. The similarity score is computed as the average cosine similarity between the generated level and a set of human levels $D_H^T$ in a task-wise manner:
\[
\mathrm{ViT}(s_{T}) = \frac{1}{|D_H^T|} \sum_{s_{\mathrm{human}} \in D_H^T} \cos\left(f_{\mathrm{ViT}}(s_{\mathrm{human}}), f_{\mathrm{ViT}}(s_{T})\right)
\]
This metric reflects high-level visual resemblance as perceived by a general-purpose vision model.

\textbf{\textit{Hamming}} distance measures the behavioral diversity within a policy, capturing variations in the structural outputs generated within an instruction. It evaluates local structural fidelity by counting the number of tile-wise mismatches between a generated level and a target reference. \rev{A higher score indicates greater diversity in the generated content. However, since the Hamming metric reflects tile-level variability, high scores can also be achieved by random generation. Therefore, diversity should be interpreted together with other metrics when evaluating generation quality.}

\subsubsection{Implementations}
\label{sec:implementation}
All experiments are implemented using the PCGRL-Jax \cite{earle2024scaling} environment and executed on a NVIDIA RTX 8000 GPU.
Encoder models are trained for 100 epochs with a batch size of 128.
DRL agents are trained using Proximal Policy Optimization (PPO, \cite{schulman2017proximal}) for 10 million timesteps.
\rev{All models are trained over 5 independent runs. Evaluation is conducted on a dataset of 1,360 conditions, where for each condition, 10 levels are generated from different initializations.}
Detailed model architectures and hyperparameters are provided in Appendix \ref{sec:hyperparameters}.

\begin{table*}[!h]
{\small
\centering
\caption{\textbf{Qualitative Comparison on Text-Modal Generation Results.} While the baseline methods are trained solely with environment rewards, VIPCGRL incorporates a similarity reward from shared representation. Since CPCGRL does not support text-based inputs, we instead provide corresponding numeric conditions for comparison. The results demonstrate that VIPCGRL generates levels that are not only aligned with human design preferences such as structure similarity and local patterns but also generalize effectively across diverse text prompts.}
\label{tab:text_generation}
\newcommand{\twinimage}[2]{\raisebox{-.5\height}{\includegraphics[width=1.35cm]{#1}\includegraphics[width=1.35cm]{#2}}}

\begin{tabular}{@{}p{4.7cm}p{2.7cm}p{2.7cm}p{3.0cm}|p{2.7cm}@{}}
\toprule
Text Input & CPCGRL & IPCGRL & VIPCGRL \scriptsize{($\lambda_{\mathrm{sim}}=1.0$)} & Human \\ \midrule
\vspace{-0.6cm}\texttt{\scriptsize{a few large regions are present / the map contains multiple placed large regions}} & \twinimage{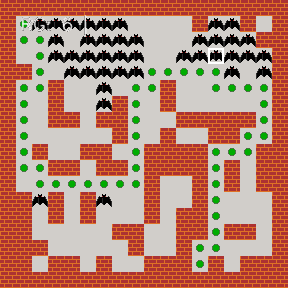}{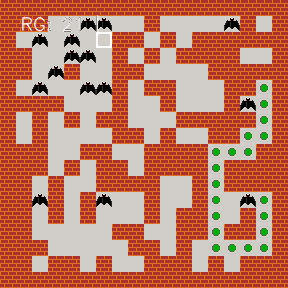} & \twinimage{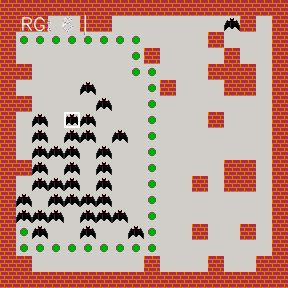}{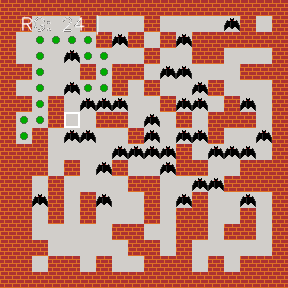} & \twinimage{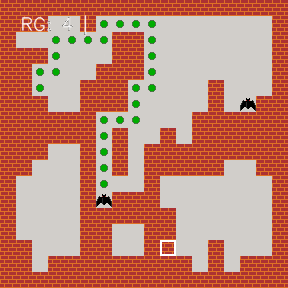}{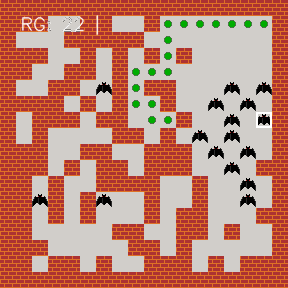} & \twinimage{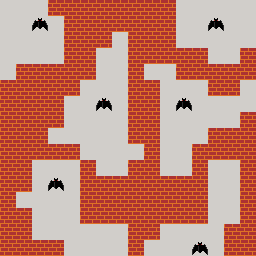}{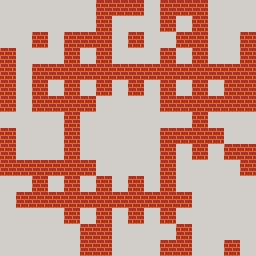} \\ 
\vspace{-0.6cm}\texttt{\scriptsize{the narrow path is extremely short and direct / long and wide path fills the map}}& \twinimage{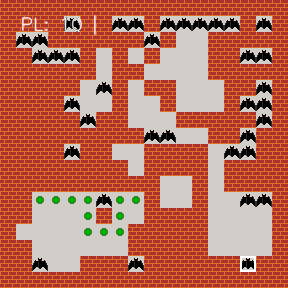}{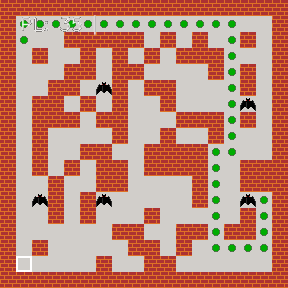} & \twinimage{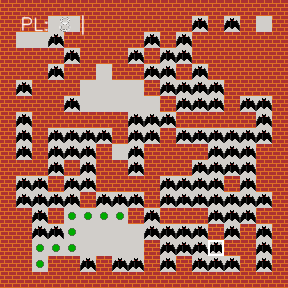}{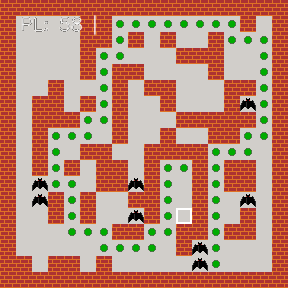} & \twinimage{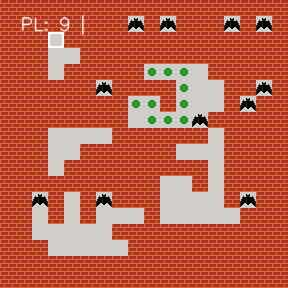}{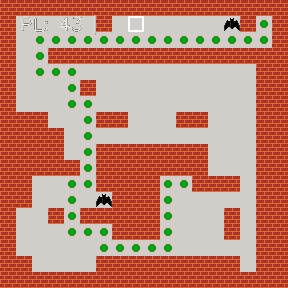} & \twinimage{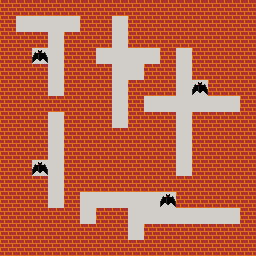}{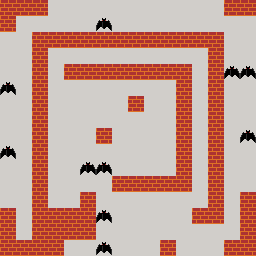} \\ 
\vspace{-0.6cm}\texttt{\scriptsize{few clustered bats are scattered / a bat swarm of scattered units emerges}} & \twinimage{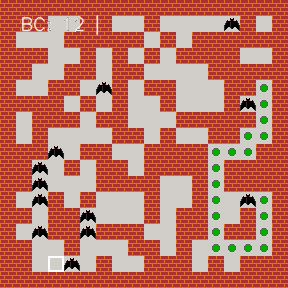}{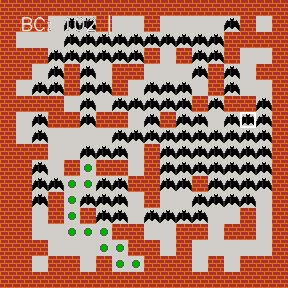} & \twinimage{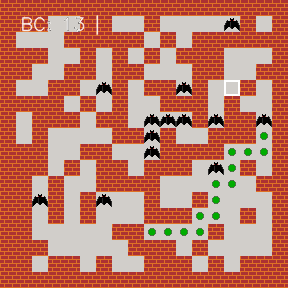}{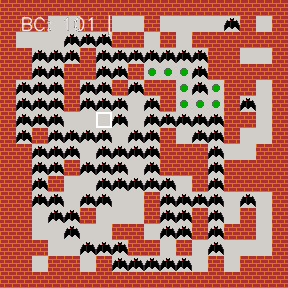} & \twinimage{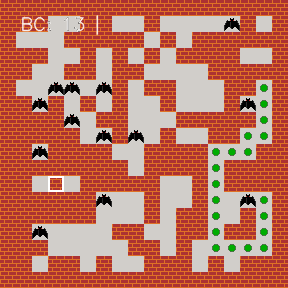}{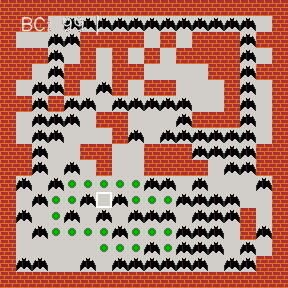} & \twinimage{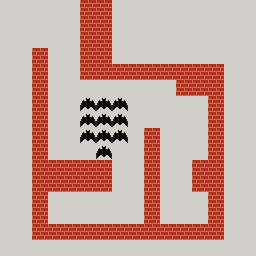}{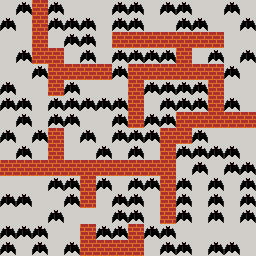} \\ \bottomrule
 \\
\end{tabular}%
}
\end{table*}

\section{Results}
\subsection{Quantitative Comparison with Previous Methods}
\label{sec:quanti_analysis}
To evaluate \textit{RH1}, we first examine whether the proposed model can achieve competitive controllability compared to prior methods while additionally enhancing human-likeness. \rev{As reported in Table~\ref{tab:quantitative_main}, VIPCGRL shows more pronounced gains in local structural similarity (TPKL-Div) and modest improvements in coarse semantic similarity (ViT-Sim) compared to IPCGRL~\cite{baek2025ipcgrl}. These improvements are statistically significant under Welch’s t-test~\cite{ruxton2006unequal} (TPKL-Div: $t=11.18$, $p<0.001$; ViT-Sim: $t=-9.19$, $p<0.001$), indicating improved alignment between the model’s generated levels and human demonstrations, particularly in local structural patterns.} Consistently, increasing the similarity reward coefficient ($\lambda_{\mathrm{sim}}$) further enhances human-likeness, indicating that stylistic alignment can be strengthened in a controlled manner.

No significant decline is observed in the Progress metric ($t=0.97$, $p=0.36$), confirming that similarity-based alignment does not hinder the agent’s ability to advance in the task. Although a slight trade-off emerges as $\lambda_{\mathrm{sim}}$ increases, the reduction in Progress is marginal, whereas the gains in human-likeness are substantial. This imbalance strongly favors our approach, showing that the agent can retain task performance while achieving markedly more human-aligned behaviors. Together, these results demonstrate that similarity-based rewards offer fine-grained control over stylistic alignment during DRL training, thereby supporting \textit{RH1}.

\subsection{Multi-modal Generation with Shared Representation} VIPCGRL demonstrates multi-modal conditional generation capabilities through its shared embedding space, addressing \textit{RH2}. As shown in Table~\ref{tab:multi moal generation}, the agent—trained solely on text inputs—successfully generalizes to unseen modalities such as level and sketch, indicating that the shared representation captures cross-modal structure. While these vision-based modalities show moderately lower controllability compared to the text modality due to the reduced semantic clarity of their condition signals, they nevertheless outperform random generation by a clear margin and yield human-likeness scores comparable to the seen modality. This contrast highlights that, despite inherent modality-dependent limitations, the shared latent space enables coherent, human-aligned, and controllable generation across diverse input types, confirming the effectiveness of our multi-modal shared representation and supporting \textit{RH2}.

\subsection{Qualitative Evaluation via User Study}
\label{sec:quali_analysis}
To practically evaluate human-likeness (\textit{RH1}), we conduct a user study.
A total of 26 human subjects participated in the evaluation, each completing a 30-minute session and providing ratings on a 7-point Likert scale. The study was approved by the institutional review board, and participants were appropriately compensated for their participation. The participants (age: 25.1$\pm$1.8) completed all evaluation items, and the study followed a within-subject design. Human-likeness was assessed using a 7-point, author-designed questionnaire, which demonstrated high internal consistency with a Cronbach’s $\alpha$ \cite{cronbach1951coefficient} of 0.89.

\rev{To evaluate human perception of the generated outputs, we design questionnaire-based criteria corresponding to each metric.}
\textbf{Human-like} measures the stylistic similarity between outputs generated by the model and human-designed datasets.
\textbf{Align-TX} assesses the model’s ability to conditionally generate outputs aligned with text-based instructions, evaluating how well the generated content satisfies the given textual conditions.
\textbf{Align-SK} evaluates the structural similarity between generated outputs and sketch-level inputs, assessing how faithfully the generation reflects the visual constraints imposed by the sketches.
Examples of text-level and sketch-level input pairs are provided in Table \ref{tab:multi moal generation}, and 20 instructions are sampled with consideration for condition balance.
We report significance using Welch’s t-test \cite{ruxton2006unequal}.

\begin{table}[t]
\small
\caption{Human evaluation results on human-likeness, and text- and sketch-alignment scores.}
\vspace{-0.1cm}
\label{tab:human_eval}
\begin{tabular}{l
    >{\centering\arraybackslash}p{1.8cm}
    >{\centering\arraybackslash}p{1.7cm}
    >{\centering\arraybackslash}p{1.7cm}}
\toprule
 &  & \multicolumn{2}{c}{Controllability} \\
 & Human-like$\uparrow$ & Align-TX$\uparrow$ & Align-SK$\uparrow$ \\
\midrule
CPCGRL & 2.81 \std{0.85} & \textminus & \textminus \\
IPCGRL & 2.90 \std{0.87} & 3.41 \std{0.91} & \textminus \\
\begin{tabular}[c]{@{}l@{}}VIPCGRL\\[-0.8ex]\scriptsize($\lambda_{\mathrm{sim}}=1.0$)\end{tabular} & \textbf{4.06} \std{0.88} & \textbf{4.25} \std{0.59} & 3.68 \std{0.89} \\
\bottomrule
\end{tabular}
\end{table}

Table \ref{tab:human_eval} summarizes the descriptive results of the user study survey.
VIPCGRL significantly outperformed the baselines in both human-likeness and text-conditional alignment.
For human-likeness, VIPCGRL scored significantly higher than CPCGRL \cite{earle2021learning} ($t=-12.37$, $p<0.001$, \rev{$d=0.77$}) and IPCGRL \cite{baek2025ipcgrl} ($t=-11.49$, $p<0.001$, \rev{$d=0.71$}).
For Align-TX, VIPCGRL also outperformed IPCGRL with statistical significance ($t=-8.87$, $p<0.001$, \rev{$d=0.55$}).
This suggests improved comprehension and execution of natural language instructions by the agent.
The Align-SK score appears to reflect the controllability characteristics of the sketch modality in Table~\ref{tab:human_eval}.
Notably, the model aligns well with the sketch modality, despite it being unseen during training. This highlights its ability to generalize across input modalities.
This trend is also consistently reflected in the user study results, indicating that the observed alignment performance translates well to human-perceived quality.

\subsection{Ablation Study on Shared Representation}
\label{sec:ablation}
We conduct an ablation study to assess whether the shared representation helps integrate multiple modalities and improves the performance of conditional DRL (\textit{RH2}).
For this, we individually train encoders by reducing the data for each specific modality (text to 25\%, level to 10\%, and sketch to 10\%) while keeping the data for the other modalities intact. 
We then use these encoders to train the DRL agent and measured the performance gap. 
The experimental results are shown in Table \ref{tab:ablation}, where each $\Delta$ value represents the performance difference between VIPCGRL and the ablated models.

Table \ref{tab:ablation} shows that reducing the data of any single modality leads to measurable changes in performance, indicating that all three data reductions influence the quality of the learned encoder representations and that each modality contributes complementarily to the shared embedding space.
Among them, the most pronounced effect is observed under the \textit{Text} data reduction setting: limiting textual data induces substantially larger performance changes in the \textit{Level} and \textit{Sketch} conditions, suggesting that text serves as a primary modality for structuring and aligning these representations during contrastive learning.
Interestingly, we further observe that data reduction can lead to a decrease in TPKL-Div value, implying that changes in data availability alter the expressive resolution of the embedding space and consequently affect the similarity-based reward signal used in agent training.
We discuss the relationship between data reduction, embedding characteristics, and reward behavior in the subsequent discussion section.

\begin{table}[ht]
\footnotesize
\caption{Results of the ablation study on data modalities. We invert the TPKL divergence so that positive $\Delta$ uniformly denotes better performance by the proposed model.}
\label{tab:ablation}
\centering
\begin{tabular}{
  >{\centering\arraybackslash}p{1.4cm} @{\hspace{15pt}} 
  l @{\hspace{22pt}} 
  c @{\hspace{22pt}} c 
}
\toprule
Data Reduction   & Input      & $\Delta$Progress$\uparrow$            & $\Delta$TPKL-Div$\downarrow$   \\
\midrule
\multirow{3}{*}{{\begin{tabular}[c]{@{}c@{}}Text\\100 $\rightarrow$ 25\%\end{tabular}}}  & Text          & -0.06 \std{0.13}     & -0.69 \std{0.89} \\
                            & Level         & -0.24 \std{0.12}     & -0.55 \std{1.01}       \\
                            & Sketch        & -0.32 \std{0.11}     & -0.48 \std{1.03}           \\
\midrule
\multirow{3}{*}{{\begin{tabular}[c]{@{}c@{}}Level\\100 $\rightarrow$ 10\%\end{tabular}}} & Text          & -0.04 \std{0.11}     & -0.30 \std{0.73}        \\
                            & Level         & -0.11 \std{0.08}     & -0.22 \std{0.72}       \\
                            & Sketch        & -0.02 \std{0.10}     & -0.31 \std{0.78}       \\
\midrule
\multirow{3}{*}{{\begin{tabular}[c]{@{}c@{}}Sketch\\100 $\rightarrow$ 10\%\end{tabular}}}& Text          & +0.02 \std{0.14}    & -0.29 \std{0.70}        \\
                            & Level         & -0.04 \std{0.16}     & -0.28 \std{0.71}   \\
                             & Sketch        & -0.09 \std{0.10}     & -0.31 \std{0.81}        \\
\bottomrule
\end{tabular}
\end{table}

\section{Discussion}
\label{sec:discussion}

\subsection{Effects of Embedding Uniformity on Similarity-Based Rewards}
We examine embedding Uniformity loss \cite{wang2020understanding} to gain insight into how efficiently the learned representations occupy the 64-dimensional latent space, where a lower uniformity value reflects a more even dispersion of embeddings and, consequently, greater representational capacity.
Across the ablation models trained with reduced modality data, we observe a tendency toward increased uniformity values (Text: +0.11$\pm{0.19}$, Level: +0.08$\pm{0.13}$, Sketch: +0.05$\pm{0.12}$), which indicate a simplification of the embedding space structure.
One possible explanation is the reduced diversity and quantity of negative samples during contrastive learning, which can lower the optimization complexity of representation learning.
Such structural simplification may limit the agent’s ability to capture fine-grained generation constraints; however, it can simultaneously enhance the model’s ability to distinguish coarse-grained level styles.
By attenuating fine-grained representational details, this reduced expressivity may discourage overly optimized or unnatural patterns and bias the learned policy toward a distribution closer to human data.
Overall, these observations suggest that adjusting the granularity of state representations could offer a practical means of balancing functional optimization and naturalness in co-creative generation scenarios.

\subsection{Non-Linear Effects of Similarity Rewards on Conditional Rewards}
The experimental results suggest that increasing the similarity reward leads to a minor trade-off with the loss associated with satisfying the generation condition, stemming from a misalignment in incorporating two types of reward. 
While the condition reward follows a linear formulation, the similarity reward, computed from embedding distances, measures proximity to the condition in a non-linear manner.
Our study highlights that visual embeddings are particularly effective in capturing coarse semantic features such as human style.
The fuzzy reward problem, which arises from non-linear semantic rewards, has been discussed in-depth in FuRL \cite{fu2024}.
To address this challenge, we propose a method that introduces a coefficient to explicitly control the balance between the two different types of reward signals.
This work contributes to the broader discourse on co-creativity by emphasizing that, in such contexts, the directional alignment of AI behavior can be more important than constraint satisfaction optimality.
\rev{A detailed task-wise analysis of the resulting reward misalignment and the feasibility of alignment across different tasks is provided in Appendix \ref{sec:misalignment}.}

\subsection{\rev{Limitations of Sketch Representation for Condition Encoding}}
\rev{We employ procedurally generated sketch images derived from game levels to ensure reproducibility and scalability; however, these sketches do not reflect sketches created by human designers in real-world settings. Through empirical analysis, we find that the effectiveness of sketch-based conditioning depends critically on the level of structural fidelity preserved in the representation. In particular, sketches that maintain explicit tile-level patterns (e.g., grid-like structures) facilitate the capture of meaningful patterns by the model, whereas overly abstract forms—such as grayscale-only images or contour-based sketches composed solely of lines—fail to adequately encode spatial and semantic distinctions between game elements. These observations suggest that specific visual characteristics are required for effectively capturing generation conditions, and future work could explore broader ranges of sketch styles while preserving these structural properties for reliable representation learning.}

\section{Conclusion and Future Work}
This paper presents VIPCGRL, a multi-modal conditioned DRL framework for generating 2D game levels.  
The proposed quadruple contrastive learning enables not only multi-modal condition representation but also alignment between human--AI behaviors.  
In addition, we introduce new human-expressive modalities, level and sketch, extending the control modality of the generative model.  
Experimental results show that the aligned DRL policy achieves higher human-likeness scores, with only a few trade-off in conditional optimality.  
Both quantitative metrics and human evaluations further support the effectiveness of the proposed method.

A limitation of VIPCGRL lies in its two-step training setup, where the pretrained contrastive encoders are frozen during the DRL stage. This design prevents the embedding space from adapting to unseen or intermediate level states encountered by the agent, potentially limiting generalization and responsiveness during policy exploration. Future work could address this by adopting an end-to-end training scheme, allowing the encoder to be fine-tuned jointly with the policy for improved adaptability, representation quality, and overall robustness.

\section{Acknowledgment}
This work was supported by the National Research Foundation of Korea (NRF) grant funded by the Korea government (MSIT) (RS-2026-25520248).
This research was supported by Basic Science Research Program through the National Research Foundation of Korea (NRF) funded by the Ministry of Education (RS-2024-00395665).
This work was supported by Institute of Information \& communications Technology Planning \& Evaluation (IITP) grant funded by the Korea government (MSIT) (No.2019-0-01842, Artificial Intelligence Graduate School Program (GIST)).
This research was supported by the ‘Project for science and technology opens the future of the region’ program, funded by the Ministry of Science and ICT(MSIT) and Gwangju Metropolitan City, Republic of Korea in 2026 (Project Name:Convergence culture virtual studio for realizing artificial intelligence based metaverse).
We appreciate the high-performance GPU computing support of HPC-AI Open Infrastructure via GIST SCENT.

\section*{Appendix}
\label{sec:appendix}

\section{2D Level Generation Environment}

The 2D level generation problem has long been studied in the field of procedural content generation (PCG), with approaches ranging from search-based methods to generative models such as GANs and diffusion models. More recently, DRL has also been explored for this task, offering a promising framework for co-creative systems due to its strength in sequential decision-making. The environment used in this study is based on the PCGRL framework introduced by \cite{khalifa2020pcgrl}, which simplifies the dungeon levels from \textit{The Legend of Zelda} (Nintendo, 1986), and has since been extended to 2D controllable PCGRL \cite{earle2021learning}, 3D level generation \cite{jiang2022learning}, and instruction-conditioned PCGRL \cite{baek2025ipcgrl}. We adopt a JAX-based GPU-accelerated implementation of the PCGRL framework\footnote{\url{https://github.com/smearle/pcgrl-jax}}, which has been reported to be up to 17 times faster than the original environment.

\subsection{Representation}
The environment offers a Gym-like representation following a Markov Decision Process.

\textbf{Observation Space}
The environment consists of an $N^2$ grid-based map ($N=16$ in this work), where each grid cell can be one of three types: \texttt{Empty} (a traversable tile that the agent can move freely across), \texttt{Wall} (an impassable and indestructible obstacle), or \texttt{Bat} (an impassable obstacle that interferes with movement). Each episode begins with a map randomly initialized using these tile types based on a weighted sampling function. The agent has full observability of the level, represented as a $(16, 16, 4)$ tensor, where the final dimension encodes the three tile types and the agent's location.

\textbf{Action Space}
The PCGRL environment supports three types of action space representations: \textit{Narrow}, \textit{Turtle}, and \textit{Wide}.

The \textit{Turtle} representation enables the agent to traverse the map, moving step by step while editing the tile at its current position.
This representation also requires longer episode lengths, since the agent must explicitly determine both movement and editing actions. The episode length of the \textit{Turtle} representation is 1,500.
The agent’s action space under the \textit{Turtle} representation comprises seven discrete actions: four movement actions (up, down, left, right), which shift the agent’s position by one tile, and three editing actions that modify the current tile to either \texttt{Empty}, \texttt{Wall}, or \texttt{Bat}.

\subsection{Task and Reward}

There are five task types available in the environment: (1) number of regions, (2) path length, (3) wall distribution, (4) monster distribution, and (5) monster direction. The first four tasks were introduced in Conditional PCGRL \cite{earle2021learning}, while the final task was proposed in Instructed PCGRL \cite{baek2024chatpcg} to investigate whether a language model can learn to satisfy spatial constraints.
Each task defines a metric function $f_{\text{metric}}$ that computes a scalar property of the current map. The agent receives a control reward $r^{\text{ctrl}}$ based on the degree of alignment between the computed metric and a predefined target value $v_{\text{goal}}$.

\textbf{Number of Regions}
This task encourages the generation of maps with a desired number of disconnected traversable regions. A region is defined as a group of connected empty tiles that are completely enclosed by \texttt{Wall}, \texttt{Bat}, or the map boundary, effectively segmenting the map into isolated areas. To identify such regions, a flood fill algorithm is applied to the empty tiles. The corresponding reward function computes the reduction in loss between the previous and current maps with respect to the target number of regions, using a region-based loss metric.

\textbf{Path Length}
This task encourages levels with a specified navigational complexity. The agent is rewarded for steering the maximum geodesic path length between two passable tiles (empty spaces) toward a target value. To estimate this metric, we run flood‑fill (BFS) from multiple randomly sampled empty tiles and record the longest valid path length observed across all source–destination pairs. The reward increases as the measured maximum path length approaches the target.

\textbf{Wall Distribution}
This task controls the overall quantity of wall tiles in the generated map, influencing the spatial complexity of the layout. The reward is based on the change in absolute loss between the desired and actual wall counts. This guides the agent toward creating either open spaces with minimal obstructions or intricate, densely partitioned structures.
The agent receives higher rewards as the number of wall tiles approaches the specified target.

\textbf{Monster Distribution}
This task controls the number of enemy tiles (\texttt{Bat}), which act as enemy-like entities in the environment. The reward is computed by measuring the reduction in absolute loss between the desired and actual bat counts before and after the agent's action.
The agent receives higher rewards as the number of enemy tiles approaches the specified target.

\textbf{Monster Direction}
This task controls the directional bias in the placement of the enemy tiles (\texttt{Bat}).
The environment is conceptually divided into horizontal and vertical halves, and the reward is computed based on the agent’s ability to match the intended directional asymmetry.
The reward penalizes deviation in the number of bats from the target region while softly penalizing placement in non-target areas.

\newlength{\tableimglen}
\setlength{\tableimglen}{4cm}

\newcommand{\tripleimage}[3]{%
  \raisebox{-.5\height}{%
    \includegraphics[width=0.33\tableimglen]{#1}\hspace{0.03cm}%
    \includegraphics[width=0.33\tableimglen]{#2}\hspace{0.03cm}%
    \includegraphics[width=0.33\tableimglen]{#3}%
  }%
}

\section{Sketch Translation Function}

\begin{figure*}
    \centering
    \includegraphics[width=1.0\linewidth]{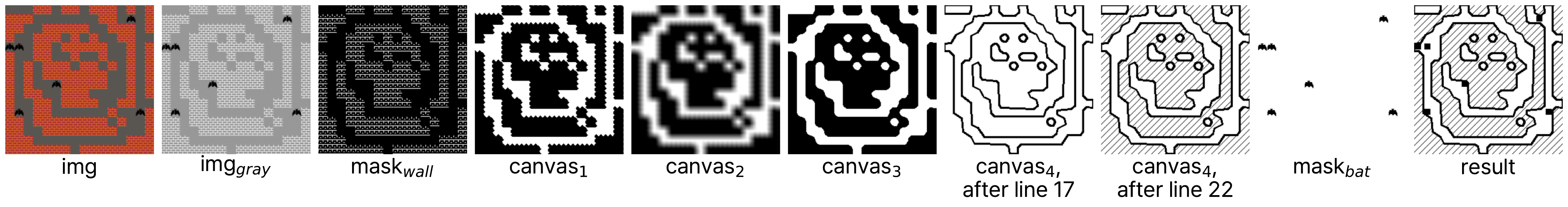}
    \vspace{-0.6cm}
    \caption{\textbf{Sketch Image Translation Process.} Starting from a color level map generated by OpenCV functions, the pipeline progressively transforms the level image into a sketch-like representation while preserving the overall level structure. 
    }
    \label{fig:sketch_translater_process}
\end{figure*}

The sketch data generation process was conducted in an OpenCV environment and can be divided into three main stages. The outline of the algorithm is presented in Algorithm~\ref{alg:sketch-translation}, and the stepwise image transformation process can be found in Fig.~\ref{fig:sketch_translater_process}.

\begin{figure}
    \centering
    \includegraphics[width=\linewidth]{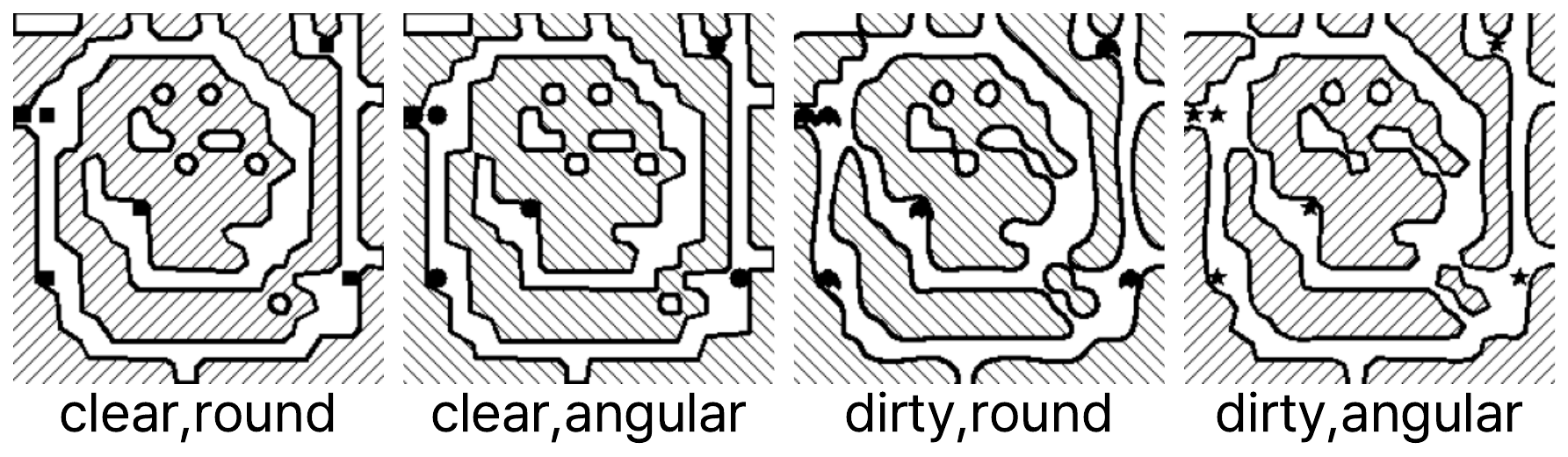}
    \vspace{-0.5cm}
    \caption{\textbf{Sketch Images Generated in Various Styles.} The same level is converted into each sketch style. Depending on the contour interpolation method, the results are categorized as either clear or dirty, and each can be further refined into round or angular substyles by adjusting the parameters. 
    }
    \label{fig:sketch_translater_style}
\end{figure}

\subsection{Pre-processing}

The inputs to the process are the level image and three style parameters: \textit{sketch style}, \textit{bat style}, and \textit{wall style}.  
If not explicitly specified, each style is randomly selected from valid candidates.

\textbf{Sketch Style} specifies the sketching method for distinguishing wall and empty regions. Two main types are supported (\textit{clear}, \textit{dirty}), which differ in their approach to contour interpolation. Each style can be further subdivided into two sub-styles (\textit{round}, \textit{angular}) by tuning their respective parameters. Examples of each style are shown in Fig.~\ref{fig:sketch_translater_style}.

\textbf{Bat Style} determines the visual representation of bats in the sketch. Available options include \textit{circle}, \textit{square}, \textit{star}, and \textit{original} (using the original bat shape).

\textbf{Wall Style} sets the direction of slashes used for wall regions, such as \textit{left slash} or \textit{right slash}.

The main parameters, $\theta$, $b_r$, $p$, and $s$, depend on the selected \textit{sketch Style}. Here, $\theta$ controls the sensitivity of wall detection; setting a higher value for $\theta$ increases the proportion of regions recognized as wall. The parameter $b_r$ governs the extent of Gaussian blurring applied during preprocessing; as $b_r$ increases, the resulting sketch becomes more rounded due to greater smoothing. The parameters $p$ and $s$ are used exclusively for the dirty style, where they influence the spline interpolation: $p$ sets the number of interpolation points, thereby controlling the level of detail in the curve, while $s$ adjusts how closely the spline follows the original contour, with higher $s$ producing smoother and more rounded shapes.

Given an input level image and selected style/parameters, the preprocessing proceeds as follows: (1) The input image is resized to $(224, 224, 1)$ and converted to grayscale to obtain $img_{gray}$. (2) To detect wall regions, thresholding with parameter $\theta$ is applied; pixels with intensity above $\theta$ are considered wall, others as empty, resulting in $mask_{wall}$. (3) The boundaries of these detected wall regions are extracted, and thick lines are drawn along the boundaries to create $canvas_1$. This approach effectively fills the wall regions and produces rough, irregular boundaries. (4) Gaussian blurring, controlled by parameter $b_r$, is applied to $canvas_1$ to smooth out irregular boundaries and generate $canvas_2$. (5) Finally, thresholding with $\theta$ is applied again to $canvas_2$ to produce the final wall mask, $canvas_3$.

\subsection{Contour Interpolation (Sketching)}

\textit{Interpolation} is a mathematical technique for smoothly connecting given data points, generating continuous curves or values between them.  
In this process, we intentionally introduce errors during interpolation to create sketch images that contain controlled distortions while still reflecting the structural properties of the original map.
Depending on the selected style, two different interpolation methods are used:

\textbf{Clear Style} uses OpenCV's \texttt{approxPolyDP} algorithm to approximate polygonal curves. By specifying an error tolerance (epsilon), the algorithm interpolates the contour into a polyline with the minimum number of vertices within the allowed error. A larger epsilon yields a more angular and simplified shape, while a smaller epsilon produces a closer match to the original. In our implementation, we varied the degree of Gaussian blurring (rather than epsilon) to distinguish more rounded or more angular styles.

\textbf{Dirty Style} uses spline interpolation via Scipy's \texttt{splprep} function. This method connects the contour points with a periodic spline curve, controlled by two parameters: the number of interpolation points $p$ and the smoothing parameter $s$. Higher $p$ provides a more finely sampled and smoother curve, while a higher $s$ produces more rounded and simplified contours. By tuning $p$ and $s$, we can generate a spectrum of sketch styles from highly rounded to more angular. 

The resulting sketch is saved as $canvas_4$.

\subsection{Region Segmentation and Post-processing}

After interpolation, slight localization errors may occur in the wall regions. Thus, instead of simply coloring the original wall positions, we segment the interpolated image into regions and determine for each region, whether it should be treated as a wall.
We use OpenCV's \texttt{connectedComponents} function to segment the image, which labels contiguous regions of identical color. Since the post-interpolation sketch image has black lines for contours and white for interior regions, this function clearly labels all regions and their boundaries.

Each region is then compared with $canvas_3$ to determine whether it represents a wall, by checking if the majority of pixels within the region correspond to wall areas (i.e., black pixels) in $canvas_3$. 
Regions classified as wall are rendered using the specified wall style (e.g., left or right slash), while boundary pixels are always rendered with a black outline.

Finally, bats are detected and rendered.  
Bat positions are identified using a low threshold on $img_{gray}$, and each detected bat is drawn on the image using the specified bat style.
Through this process, we generate sketch images that maintain the essential map features and clear tile states, while allowing controlled error and stylistic diversity. For dataset construction, each image was generated using randomly selected styles and parameters to enhance variation.

\begin{algorithm}[]
\caption{Sketch Translation Algorithm}
\label{alg:sketch-translation}
\begin{algorithmic}[1]
\STATE \textbf{Input:} $img$, style parameters $\mathit{(sketch,~bat,~wall)}$
\STATE $(\theta,\, b_r,\,\, p,\, s)$ according to $\mathit{sketch\_style}$
\STATE $img_{small} \gets$ ReSize($img$, $224\times224$) 
\STATE $img_{gray} \gets$ GrayScale($img_{small}$)
\STATE $mask_{wall} \gets$ Threshold($img_{gray}$, $\theta$)
\STATE $canvas_1 \gets$ Draw(FindContours($mask_{wall}$), style:thickline)
\STATE $canvas_2 \gets$ GaussianBlur($canvas_1$, $b_r$)
\STATE $canvas_3 \gets$ Threshold($canvas_2$, $\theta$)
\STATE $wallContours \gets$ FindContours($canvas_3$)
\IF{$sketch\_style$ = ``clear''}
    \FOR{$cnt$ in $wallContours$}
        \STATE $canvas_4 \gets$ Draw(approxPolyDP($cnt$))
    \ENDFOR
\ELSIF{$sketch\_style$ =``dirty''}
    \FOR{$cnt$ in $wallContours$}
        \STATE $canvas_4 \gets$ Draw(SplineSmoothing($cnt$, $p$, $s$))
    \ENDFOR
\ENDIF
\FOR{region in Components($canvas_4$)}
    \IF{MajorColor($canvas_3$, region) = black}
        \STATE $canvas_4 \gets$ Draw(region, style: $wall\_style$)
    \ENDIF
\ENDFOR
\STATE $mask_{bat} \gets$ Threshold($img_{gray}$, $60$)
\STATE $batContours \gets$ FindContours($mask_{bat}$)
\FOR{$bat$ in $batContours$}
    \STATE $canvas_4 \gets$ Draw($bat$, style: $bat\_style$)
\ENDFOR
\STATE \textbf{Return} $canvas_4$
\end{algorithmic}
\end{algorithm}

\section{Quadruple Contrastive Loss}
We provide the full equations for the quadruple contrastive loss, $\mathcal{L}$, as introduced in the main text. The total loss $\mathcal{L}$ is the sum of six directional loss terms, covering all six directional pairs among the three modalities: Text, Level, and Sketch. The directional loss functions are defined individually below.

\begin{equation}
\begin{aligned}
\mathcal{L}_{\text{Total}} =\;& 
\mathcal{L}_{\text{Text}\to\text{Level}} 
+ \mathcal{L}_{\text{Text}\to\text{Sketch}}
+ \mathcal{L}_{\text{Level}\to\text{Text}} \\
& + \mathcal{L}_{\text{Level}\to\text{Sketch}}
+ \mathcal{L}_{\text{Sketch}\to\text{Text}}
+ \mathcal{L}_{\text{Sketch}\to\text{Level}}
\end{aligned}
\end{equation}

The notation used here specifies the general terms from the main text: the embedding vectors $z, g, \text{and } h$ correspond to the modalities text, level, and sketch ($e^{(\text{text})}, e^{(\text{level})}, \text{and } e^{(\text{sketch})}$), respectively.
\\
    \begin{equation*}
    \mathcal{L}_{\text{X} \to \text{Y}} = -\frac{1}{|B|} \sum_{i \in B} \log \frac{ \sum_{j \in P_{\text{X}\to\text{Y}}(i)} \exp\left( \frac{ \mathrm{sim}(z_i, g_j) }{\tau} \right) }{ \sum_{k \in B} \exp\left( \frac{ \mathrm{sim}(z_i, g_k) }{\tau} \right) }
    \end{equation*}

\subsection{Notation}
\begin{itemize}
    \item $B$: A minibatch.
    \item $i, j, k$: Indices of samples within the minibatch.
    \item $z, g, h$: Embedding vectors for the text, level, and sketch modalities, respectively. For example, $z_i$ is the embedding of the $i$-th text sample.
    \item $\mathrm{sim}(\cdot, \cdot)$: The cosine similarity function, as described in the main text.
    \item $\tau$: A learnable temperature hyperparameter.
    \item $P_{\text{Source}\to\text{Target}}(i)$: The set of indices of positive samples, corresponding to the generalized set $P_{m \to n}(i)$ in the main text. For example, $P_{\text{Text}\to\text{Level}}(i)$ is the set of indices for samples in the Level modality that share the same structural condition and creation style (human and AI) as the anchor sample $z_i$ from the Text modality.
\end{itemize}

In each loss term, the numerator sums the similarity scores between an anchor embedding (e.g., $z_i$) and all of its corresponding positive embeddings in the target modality. The denominator normalizes this value by the sum of similarities between the anchor and all embeddings of the target modality in the minibatch (both positive and negative). Minimizing this loss trains the encoders to map semantically related samples from different modalities to nearby points in the shared latent space, while pushing apart unrelated samples.

\section{Model Hyperparameters}
\label{sec:hyperparameters}
\subsection{CLIP Parameters}

Table~\ref{tab:clip_hyper} shows the encoder training hyperparameters used in this work and  IPCGRL~\cite{baek2025ipcgrl}.
The most of parameters are shared across IPCGRL and the VIPCGRL framework for a fair comparison in embedding-based instruction grounding.

\begin{table}
\centering
\small
\caption{Encoder training hyperparameters}
\label{tab:clip_hyper}
\begin{tabular}{p{3cm}p{1.2cm}p{1.5cm}}
\toprule
\textbf{} & \textbf{Notation} & \textbf{Value} \\
\midrule
Learning rate &  & $1 \times 10^{-3}$ \\
Batch size & $B$ & 128 \\
Number of epochs & $E$ & 100 \\
Weight decay & $c_{\text{wd}}$ & $1 \times 10^{-5}$ \\
Text encoder backbone (IPCGRL) &  & BERT \\
Text encoder backbone (VIPCGRL) &  & CLIP \\
\bottomrule
\end{tabular}
\end{table}

\begin{figure*}[]
    \centering
    \includegraphics[width=0.7\linewidth]{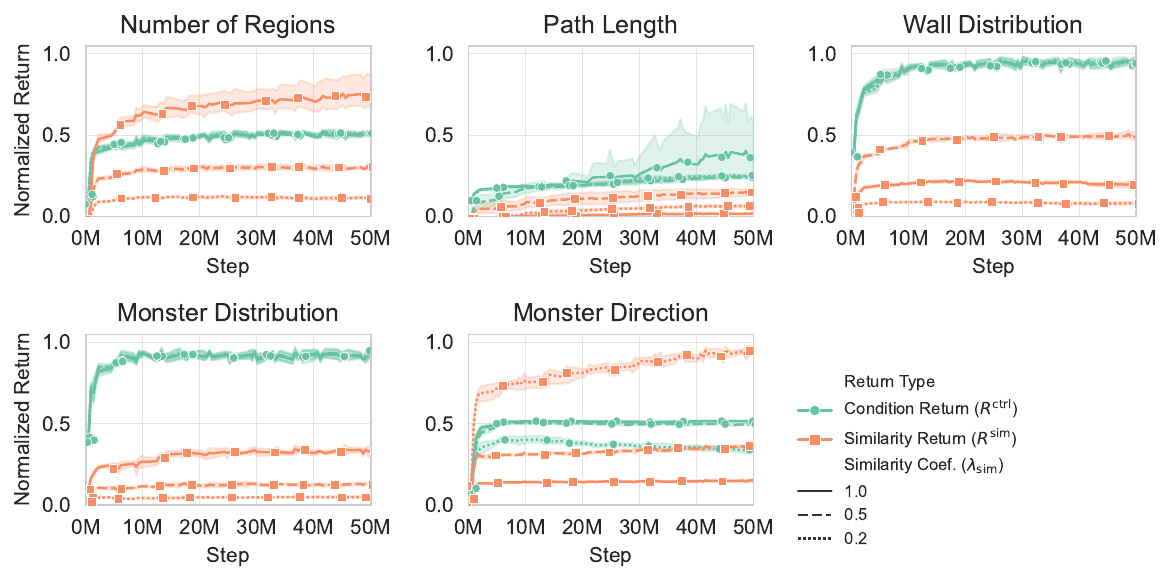}
    \vspace{-0.5cm}
    \caption{
    \textbf{Effect of Similarity Coefficient on Environment Rewards.}
    Normalized returns are shown over the course of training for various environment attributes. Each plot compares two types of return signals—the condition return ($R^{\mathrm{ctrl}}$) and the similarity return ($R^{\mathrm{sim}}$)—evaluated under different similarity coefficients ($\lambda_{\mathrm{sim}} \in \{1.0, 0.5, 0.2\}$). Shaded areas denote standard deviation across multiple runs.
    }
    \label{fig:cond_sim_reward}
\end{figure*}

The encoder processes inputs from three modalities—text, level state, and sketch image—and projects each into a 64-dimensional embedding used for conditioning the policy. For the text input in IPCGRL, tokenized instructions are first encoded using a frozen backbone model, followed by a two-layer MLP with hidden sizes of 256 and 64, where a Tanh non-linear activation is applied between the layers to map the final BERT embedding into a compact textual representation.
In the VIPCGRL setting, a single-layer MLP is employed to obtain a compact text representation.
3The backbone models use frozen pretrained BERT and CLIP models in IPCGRL and VIPCGRL, respectively.
For the level and sketch modalities, visual tokens (e.g., grid-based representation or grayscale image) are processed via modality-specific CNN encoders with residual blocks. The level encoder is relatively shallow, with two residual blocks and a single downsampling via strided convolution, while the sketch encoder has a deeper structure with four residual blocks and four downsampling stages. Both encoders apply global average pooling followed by a projection MLP to obtain 64-dimensional embeddings.
All three modality-specific embedding heads are fixed during conditional DRL; gradients are not propagated into the encoder. This design allows the encoder to provide stable, pretrained representations for language and visual conditioning without requiring end-to-end joint optimization.

\subsection{Conditional DRL Parameters}
Table~\ref{tab:rl_hyper} lists the core DRL hyperparameters used in both CPCGRL~\cite{earle2021learning}, IPCGRL~\cite{baek2025ipcgrl}, and VIPCGRL. We adopt this shared configuration to ensure consistency in the policy optimization process across methods.

\begin{table}
\centering
\caption{Conditional reinforcement learning hyperparameters}
\label{tab:rl_hyper}
\begin{tabular}{p{4cm}p{1cm}p{1.5cm}}
\toprule
\textbf{} & \textbf{Notation} & \textbf{Value} \\
\midrule
Learning rate &  & $1 \times 10^{-4}$ \\
Rollout steps before update & $n_{\text{steps}}$ & 128 \\
Total timesteps & $T$ & $5 \times 10^7$ \\
Discount factor & $\gamma$ & 0.99 \\
GAE smoothing parameter & $\lambda$ & 0.95 \\
PPO clipping threshold & $\epsilon$ & 0.2 \\
Max grad normalization & $\|\nabla\|_{\text{max}}$ & 0.5 \\
Number of parallel environments & $N_\text{env}$ & 500 \\
\midrule
Similarity return coefficient & $\lambda_{\mathrm{sim}}$ & 1.0 \\
Reward scaling factor & $\alpha$ & 30.0 \\
\bottomrule
\end{tabular}
\end{table}

The policy network adopts an actor-critic architecture and receives as input the frozen text, image, or sketch embeddings alongside environmental observations. The spatial observation map is processed through two CNN layers with kernel size 7×7, stride 2, and padding 3, each followed by an activation function (ReLU and Tanh). The output is flattened and concatenated with the scalar observation vector and all three modality embeddings. This compressed representation is passed through two-layer MLP layers with hidden dimensions 256 and 128, with using non-linear activation (Tanh). The actor head outputs a flat vector of logits, where each location selects one of 7 discrete tile-editing actions.

\section{Vision Transformer-based Human-likeness Metric}
\begin{figure}
    \centering
    \includegraphics[width=0.8\linewidth]{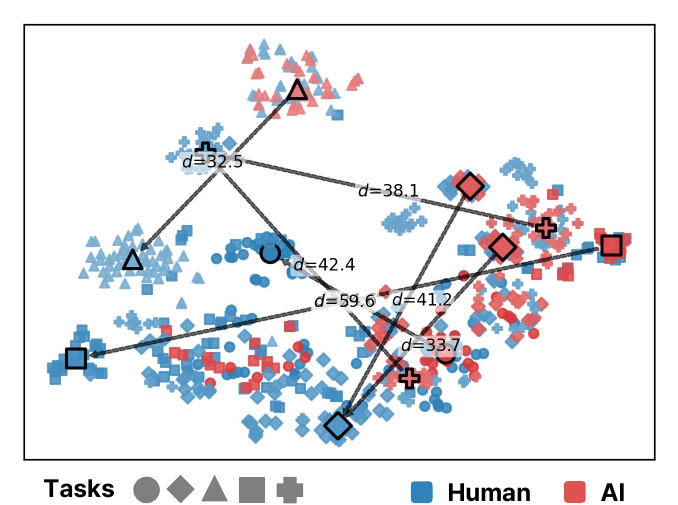}
    \caption{\textbf{Human–AI Embedding Space Separation Using a Pretrained ViT.} A t-SNE \cite{maaten2008visualizing} projection of the learned embedding space illustrates that human-authored and AI-generated levels form clearly separable clusters. The pretrained vision transformer embedding effectively captures and separates stylistic patterns in game levels.}
    \label{fig:vit_embedding}
\end{figure}

The \textit{ViT-Sim} metric is used as a measure of human-likeness, aiming to quantify the semantic similarity between generated levels and human-designed ones. Because the ViT backbone is pretrained on large-scale natural image datasets spanning diverse domains—rather than on game-level imagery—it may exhibit out-of-distribution behavior when applied to this task. Thus, it is necessary to verify whether the pretrained model can meaningfully encode the stylistic properties of game levels.
To assess this, we analyze the embedding space produced by the pretrained ViT. Fig. \ref{fig:vit_embedding} shows a t-SNE projection of embeddings for human-authored and AI-generated (IPCGRL \cite{baek2025ipcgrl}) levels across tasks. The resulting distributions reveal a clear stylistic separation: human levels occupy a broader and more diverse region of the latent space, while AI levels cluster into more compact areas. This indicates that, despite being pretrained on non-game domains, the ViT embedding robustly captures style-related differences in level layouts. % Consequently, \textit{ViT-Sim} serves as a valid metric for evaluating whether generated levels align with human design tendencies.

\section{Misalignment in Conditional and Similarity Rewards}
\label{sec:misalignment}

We analyze the effect of varying the similarity coefficient ($\lambda_{\mathrm{sim}}$) on the alignment between the similarity return ($R^{\mathrm{sim}}$) and the condition return ($R^{\mathrm{ctrl}}$). As shown in Figure~\ref{fig:cond_sim_reward}, increasing $\lambda_{\mathrm{sim}}$ consistently improves $R^{\mathrm{sim}}$, but significantly alters the trajectory of $R^{\mathrm{ctrl}}$, particularly in the \textit{Path Length} and \textit{Monster Direction} tasks, where the condition return exhibits the most noticeable departure from the baseline ($\lambda_{\mathrm{sim}}=0.2$). These results highlight the importance of carefully balancing imitation and task reward to promote alignment. The observed divergence arises in part because human demonstrations may encode task intent in ways that are not fully captured by the environment's reward. In this context, the similarity reward serves not as a conflicting signal, but as an essential component of the reward structure that helps align the policy with human-preferred behaviors. Rather than merely optimizing for externally defined objectives, the policy benefits from trajectory-level guidance encoded in human data, with the similarity objective acting as a non-linear reward function that complements and enriches environment-based learning.

\section{Multi-modal Level Generation}

\newcommand{\twoimgS}[2]{%
  \raisebox{-.5\height}{%
    \includegraphics[width=1.33cm]{#1}%
    \includegraphics[width=1.33cm]{#2}%
    }%
}

\renewcommand{\oneimg}[1]{%
  \raisebox{-.5\height}{%
    \includegraphics[width=1.33cm]{#1}%
    }%
}

Table~\ref{tab:multi_modal_generation_text_sketch} presents qualitative generation
results where text and sketch conditions are aligned per task and evaluated under
identical reward targets.
For each task, the table pairs a textual input and its generated output with the
corresponding sketch condition, sketch input, and generated output.
Across all tasks, including region count, path length, wall distribution, monster
distribution, and monster direction, the generated levels reflect the specified
conditions in both modalities, showing comparable structural patterns when conditioned
on the same target values.
These visual comparisons illustrate how different input forms lead to consistent
generation outcomes within the same task setting.

% =========================================================
% Text + Sketch Integrated Visualization Table (Full)
% =========================================================
\begin{table*}
\small
\centering
\caption{Visualization of generation results based on aligned text and sketch input modalities.}
\label{tab:multi_modal_generation_text_sketch}

\renewcommand{\arraystretch}{0.85}
\setlength{\tabcolsep}{1.7pt}
\vspace{0.3cm}

\begin{tabular}{@{}m{3cm}m{4.2cm}m{2.7cm}m{2.7cm}m{1.36cm}m{2.72cm}@{}}
\toprule
\textbf{Task} &
\multicolumn{2}{c}{\textbf{Text}} &
\multicolumn{3}{c}{\textbf{Sketch}} \\
& Input & Output & Condition & Input & Output \\
\midrule

% ===================== Number of Regions =====================
\multirow{3}{*}{\raisebox{-5em}{\textbf{Number of Regions}}}
& \textit{\textbf{A few\hidden{(5)}}} \textit{\textbf{small}} regions are present
& \twoimgS{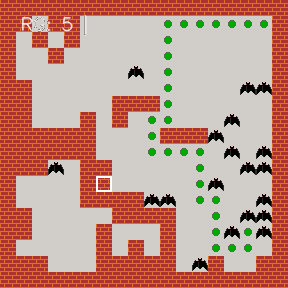}
           {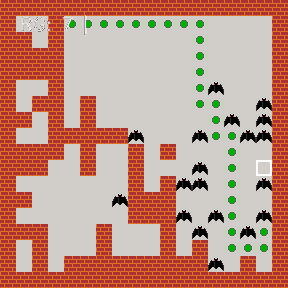}
& a few\hidden{(5)} / large
& \oneimg{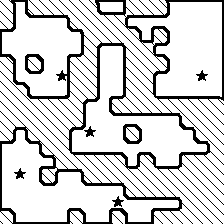}
& \twoimgS{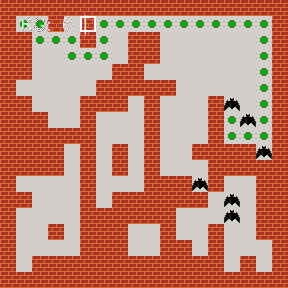}
           {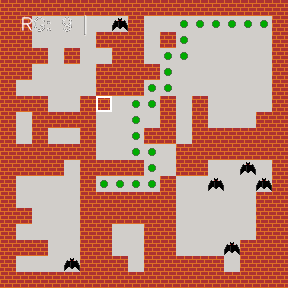}
\\

& A \textit{\textbf{moderate\hidden{(15)}}} amount of \textit{\textbf{small}} regions exist
& \twoimgS{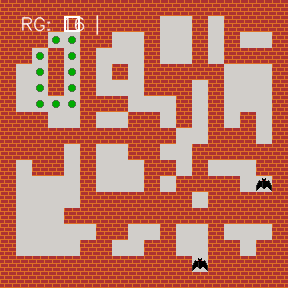}
           {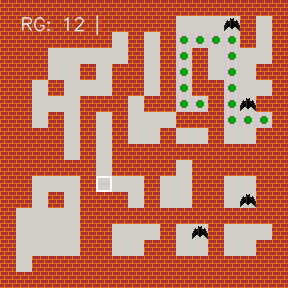}
& some\hidden{(15)} / small
& \oneimg{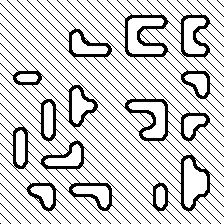}
& \twoimgS{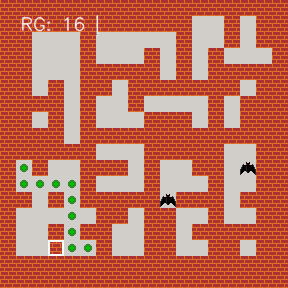}
           {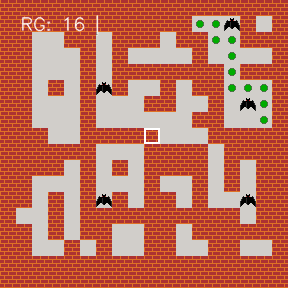}
\\

& \textit{\textbf{Several\hidden{(25)}}} \textit{\textbf{large}} regions are scattered
& \twoimgS{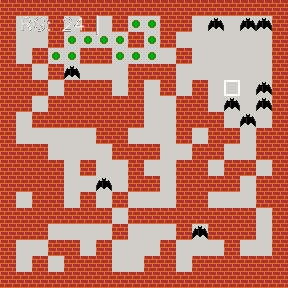}
           {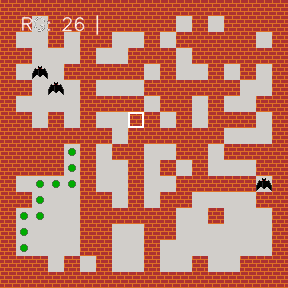}
& several\hidden{(25)} / small
& \oneimg{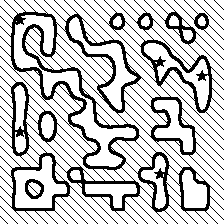}
& \twoimgS{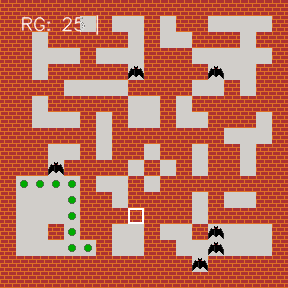}
           {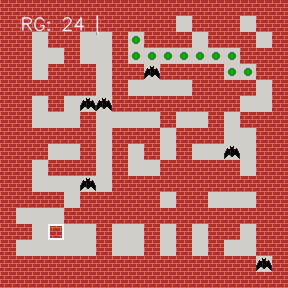}
\\

\midrule

% ===================== Path Length =====================
\multirow{3}{*}{\raisebox{-5em}{\textbf{Path Length}}}
& \textit{\textbf{Micro\hidden{(10)}}} path length using a \textit{\textbf{narrow}} corridor
& \twoimgS{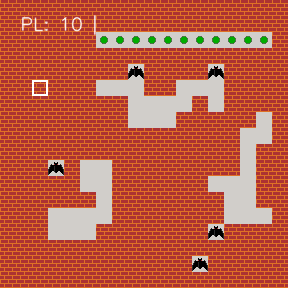}
           {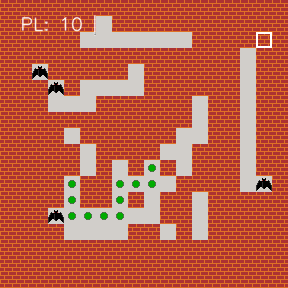}
& very short\hidden{(10)} / narrow
& \oneimg{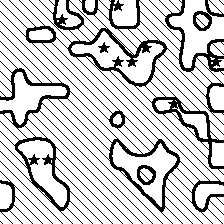}
& \twoimgS{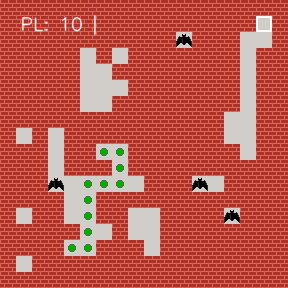}
           {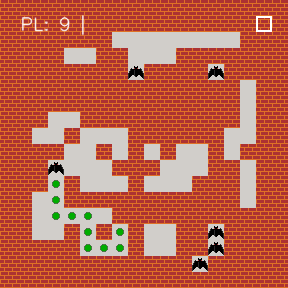}
\\

& \textit{\textbf{Short\hidden{(20)}}} path length with a \textit{\textbf{narrow}} design
& \twoimgS{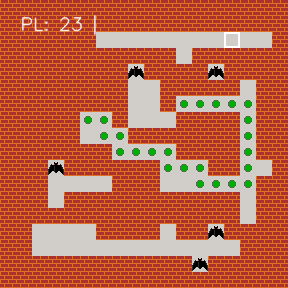}
           {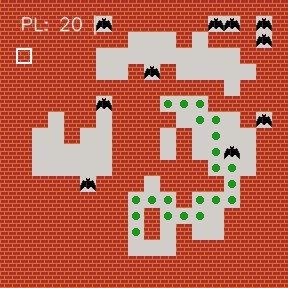}
& short\hidden{(20)} / narrow
& \oneimg{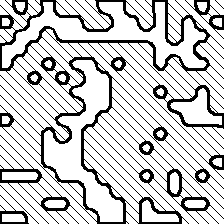}
& \twoimgS{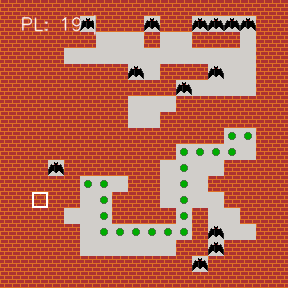}
           {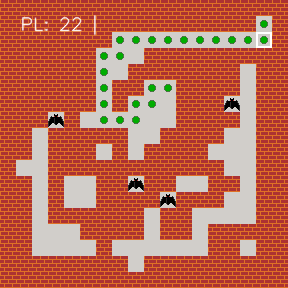}
\\

& A \textit{\textbf{balanced\hidden{(40)}}} path length with \textit{\textbf{wide}} characteristics
& \twoimgS{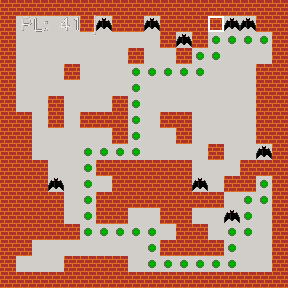}
           {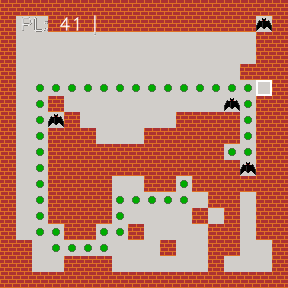}
& moderate\hidden{(40)} / narrow
& \oneimg{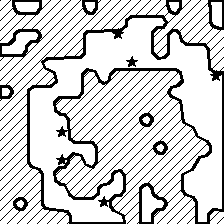}
& \twoimgS{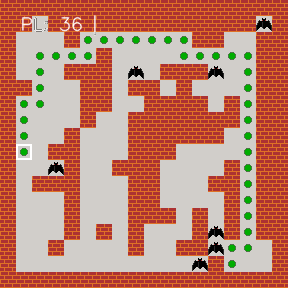}
           {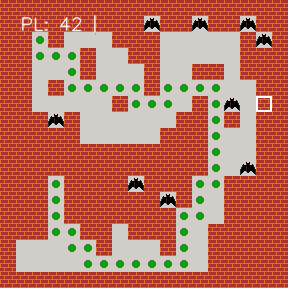}
\\

\midrule

% ===================== Wall Distribution =====================
\multirow{3}{*}{\raisebox{-5em}{\textbf{Wall Distribution}}}
& \textit{\textbf{Sparse\hidden{(40)}}} and \textit{\textbf{decentralized}} blocks occupy the map
& \twoimgS{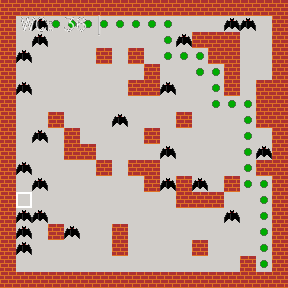}
           {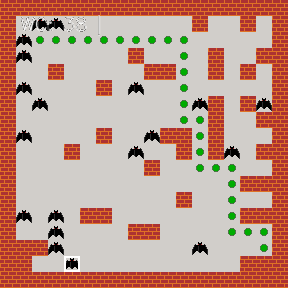}
& a few\hidden{(40)} / decentralized
& \oneimg{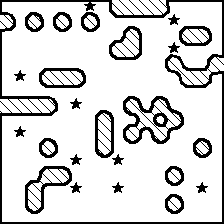}
& \twoimgS{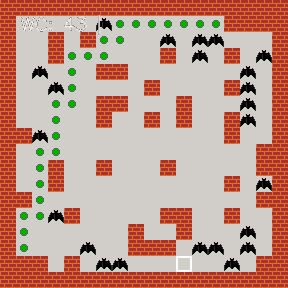}
           {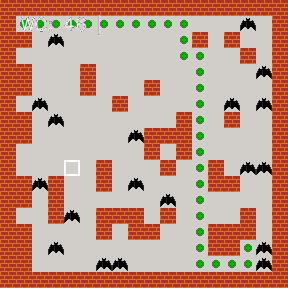}
\\

& \textit{\textbf{Centralized}} blocks appear with \textit{\textbf{moderate\hidden{(80)}}} density
& \twoimgS{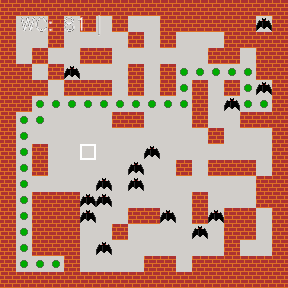}
           {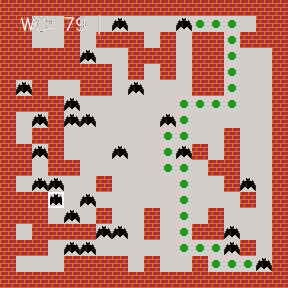}
& some\hidden{(80)} / centralized
& \oneimg{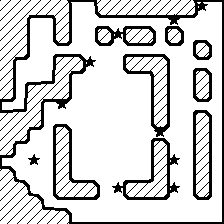}
& \twoimgS{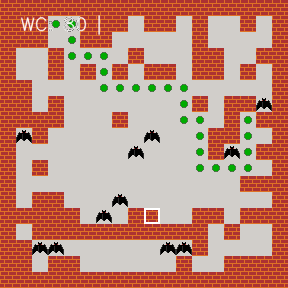}
           {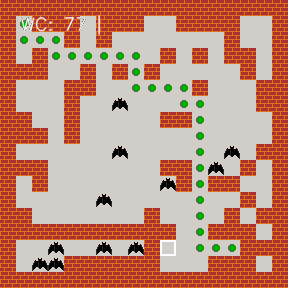}
\\

& Block \textit{\textbf{overcrowding\hidden{(120)}}} is observed in the \textit{\textbf{decentralized}} layout
& \twoimgS{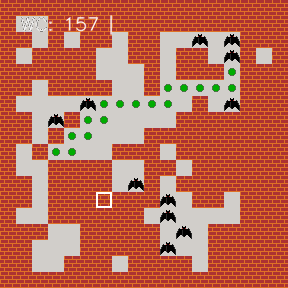}
           {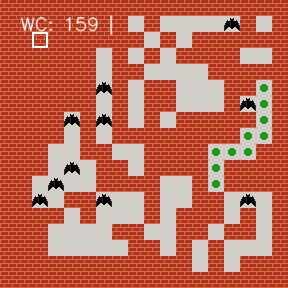}
& substantial\hidden{(120)} / centralized
& \oneimg{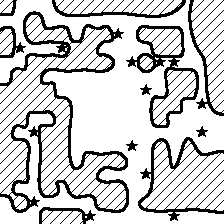}
& \twoimgS{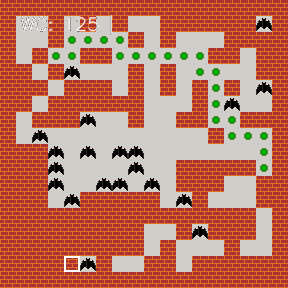}
           {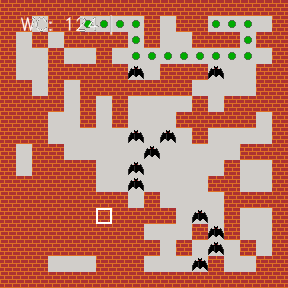}
\\

\midrule

% ===================== Monster Distribution =====================
\multirow{3}{*}{\raisebox{-5em}{\textbf{Monster Distribution}}}
& \textit{\textbf{A few\hidden{(10)} scattered}} bats appear
& \twoimgS{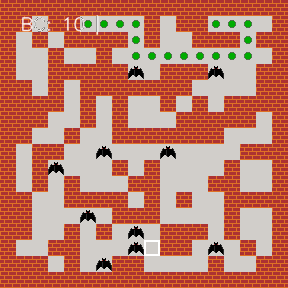}
           {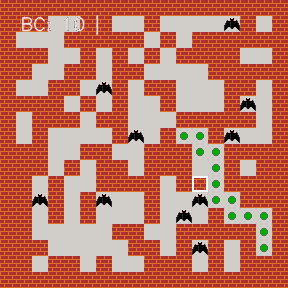}
& a few\hidden{(10)} / scattered
& \oneimg{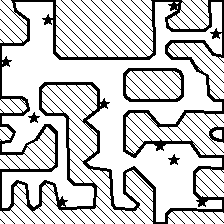}
& \twoimgS{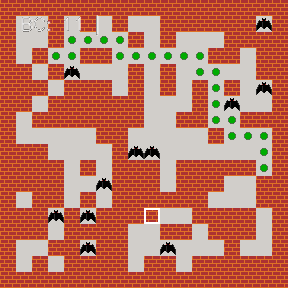}
           {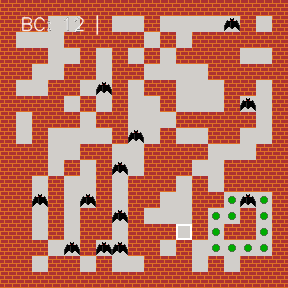}
\\

& \textit{\textbf{Some\hidden{(40)} clustered}} bats spawn throughout
& \twoimgS{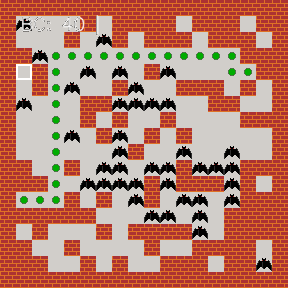}
           {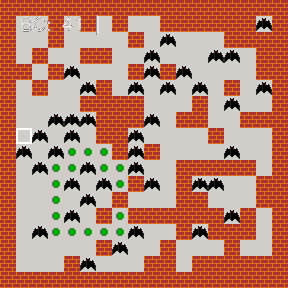}
& some\hidden{(40)} / clustered
& \oneimg{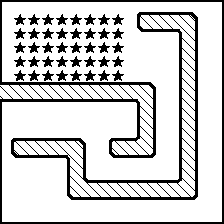}
& \twoimgS{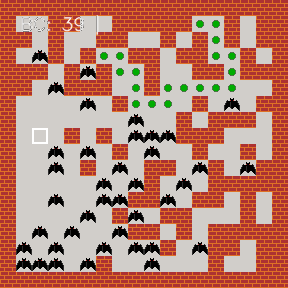}
           {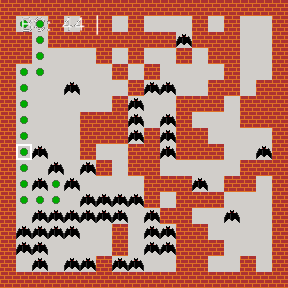}
\\

& \textit{\textbf{Many\hidden{(100)} scattered}} bats flood the map
& \twoimgS{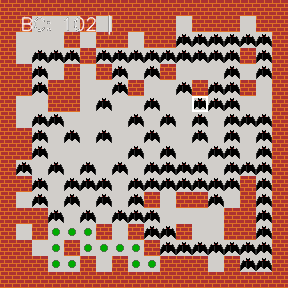}
           {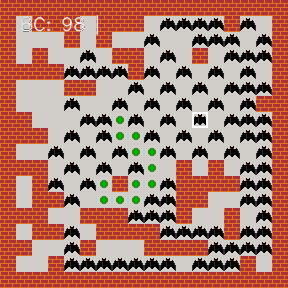}
& substantial\hidden{(70)} / clustered
& \oneimg{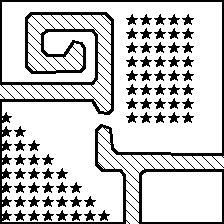}
& \twoimgS{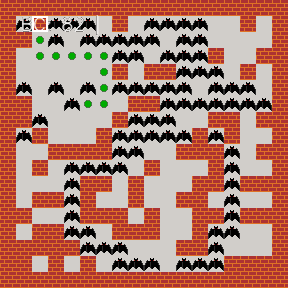}
           {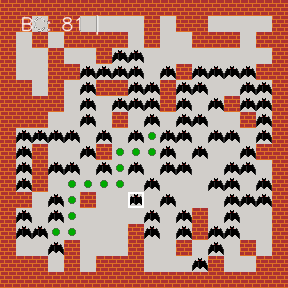}
\\

\midrule

% ===================== Monster Direction =====================
\multirow{3}{*}{\raisebox{-5em}{\textbf{Monster Direction}}}
& Bats cluster in the \textit{\textbf{west}} in a \textit{\textbf{radial}} pattern
& \twoimgS{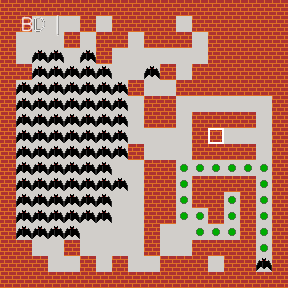}
           {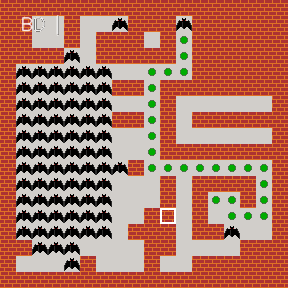}
& left / radial
& \oneimg{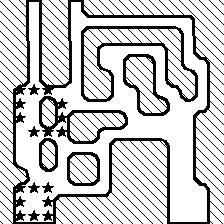}
& \twoimgS{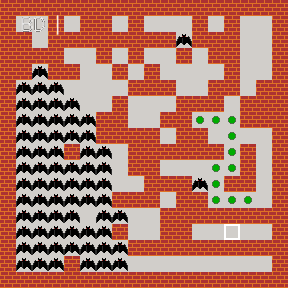}
           {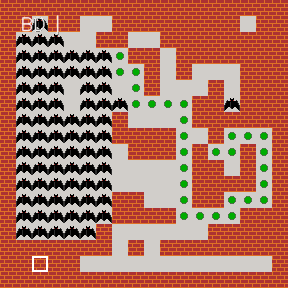}
\\

& A \textit{\textbf{radial}} bats pattern forms across the \textit{\textbf{top}}
& \twoimgS{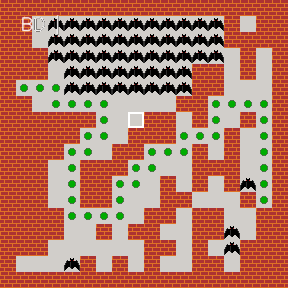}
           {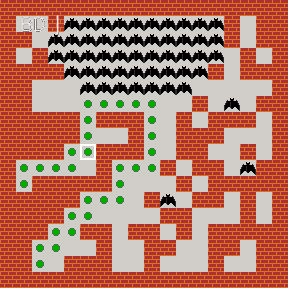}
& top / radial
& \oneimg{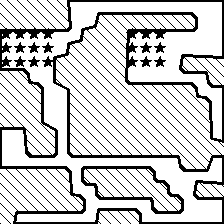}
& \twoimgS{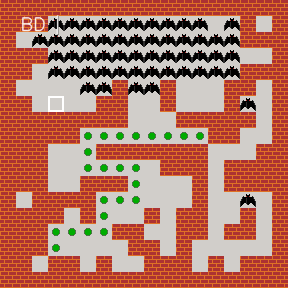}
           {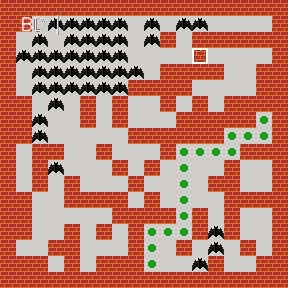}
\\

& \textit{\textbf{Radial}} bats fill the \textit{\textbf{bottom}} edge
& \twoimgS{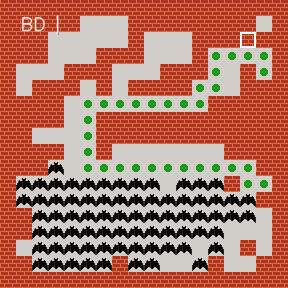}
           {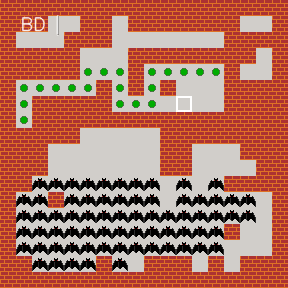}
& bottom / radial
& \oneimg{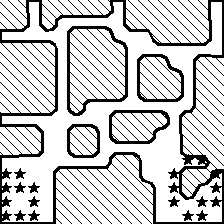}
& \twoimgS{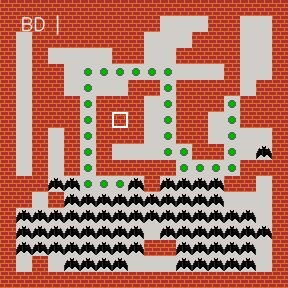}
           {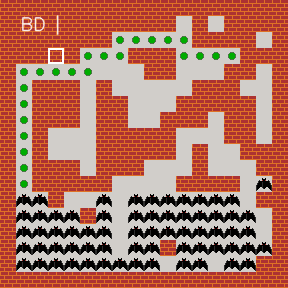}
\\

\bottomrule
\end{tabular}
\vspace{-0.3cm}
\end{table*}

\section{Dataset and Codebase}
The full implementation of our framework, along with the datasets used in our experiments, is available at \url{https://github.com/bic4907/VIPCGRL}.

\bibliographystyle{IEEEtran}
\bibliography{references}

@article{jeon2023raidenv,
  title={RaidEnv: Exploring New Challenges in Automated Content Balancing for Boss Raid Games},
  author={Jeon, Hyeon-Chang and Baek, In-Chang and Bae, Cheong-Mok and Park, Taehwa and You, Wonsang and Ha, Taegwan and Jung, Hoyoun and Noh, Jinha and Oh, Seungwon and Kim, Kyung-Joong},
  journal={IEEE Transactions on Games},
  year={2023},
  publisher={IEEE}
}

@inproceedings{khalifa2020pcgrl,
  title={Pcgrl: Procedural content generation via reinforcement learning},
  author={Khalifa, Ahmed and Bontrager, Philip and Earle, Sam and Togelius, Julian},
  booktitle={Proceedings of the AAAI Conference on Artificial Intelligence and Interactive Digital Entertainment},
  volume={16},
  number={1},
  pages={95--101},
  year={2020}
}

@inproceedings{earle2021learning,
  title={Learning Controllable Content Generators},
  author={Earle, Sam and Edwards, Maria and Khalifa, Ahmed and Bontrager, Philip and Togelius, Julian},
  booktitle={2021 IEEE Conference on Games (CoG)},
  pages={1--9},
  year={2021},
  organization={IEEE}
}

@inproceedings{jiang2022learning,
  title={Learning Controllable 3D Level Generators},
  author={Jiang, Zehua and Earle, Sam and Green, Michael and Togelius, Julian},
  booktitle={Proceedings of the 17th International Conference on the Foundations of Digital Games},
  pages={1--9},
  year={2022}
}

@article{togelius2011search,
  author    = {Togelius, Julian and Yannakakis, Georgios N. and Stanley, Kenneth O. and Browne, Cameron},
  title     = {Search-based procedural content generation: A taxonomy and survey},
  journal   = {IEEE Transactions on Computational Intelligence and AI in Games},
  volume    = {3},
  number    = {3},
  pages     = {172--186},
  year      = {2011}
}

@inproceedings{baek2024chatpcg,
  title={ChatPCG: Large Language Model-Driven Reward Design for Procedural Content Generation},
  author={Baek, In-Chang and Park, Tae-Hwa and Noh, Jin-Ha and Bae, Cheong-Mok and Kim, Kyung-Joong},
  booktitle={2024 IEEE Conference on Games (CoG)},
  pages={1--4},
  year={2024},
  organization={IEEE}
}

@inproceedings{earle2024scaling,
  title={Scaling, Control and Generalization in Reinforcement Learning Level Generators},
  author={Earle, Sam and Jiang, Zehua and Togelius, Julian},
  booktitle={2024 IEEE Conference on Games (CoG)},
  pages={1--8},
  year={2024},
  organization={IEEE}
}

@article{baek2025pcgrllm,
  title={PCGRLLM: Large Language Model-Driven Reward Design for Procedural Content Generation Reinforcement Learning},
  author={Baek, In-Chang and Kim, Sung-Hyun and Earle, Sam and Jiang, Zehua and Jin-Ha, Noh and Togelius, Julian and Kim, Kyung-Joong},
  journal={arXiv preprint arXiv:2502.10906},
  year={2025}
}

@article{liu2018intriguing,
  title={An intriguing failing of convolutional neural networks and the coordconv solution},
  author={Liu, Rosanne and Lehman, Joel and Molino, Piero and Petroski Such, Felipe and Frank, Eric and Sergeev, Alex and Yosinski, Jason},
  journal={Advances in neural information processing systems},
  volume={31},
  year={2018}
}

@article{schulman2017proximal,
  title={Proximal policy optimization algorithms},
  author={Schulman, John and Wolski, Filip and Dhariwal, Prafulla and Radford, Alec and Klimov, Oleg},
  journal={arXiv preprint arXiv:1707.06347},
  year={2017}
}

@inproceedings{radford2021learning,
  title={Learning transferable visual models from natural language supervision},
  author={Radford, Alec and Kim, Jong Wook and Hallacy, Chris and Ramesh, Aditya and Goh, Gabriel and Agarwal, Sandhini and Sastry, Girish and Askell, Amanda and Mishkin, Pamela and Clark, Jack and others},
  booktitle={International conference on machine learning},
  pages={8748--8763},
  year={2021},
  organization={PmLR}
}

@inproceedings{baek2025ipcgrl,
  title={IPCGRL: Language-Instructed Reinforcement Learning for Procedural Level Generation},
  author={Baek, In-Chang and Kim, Sung-Hyun and Lee, Seo-Young and Kim, Dong-Hyeon and Kim, Kyung-Joong},
  booktitle={2025 IEEE Conference on Games (CoG)},
  pages={1--8},
  year={2025},
  organization={IEEE}
}

@inproceedings{delarosa2021mixed,
  title={Mixed-initiative level design with rl brush},
  author={Delarosa, Omar and Dong, Hang and Ruan, Mindy and Khalifa, Ahmed and Togelius, Julian},
  booktitle={Artificial Intelligence in Music, Sound, Art and Design: 10th International Conference, EvoMUSART 2021, Held as Part of EvoStar 2021, Virtual Event, April 7--9, 2021, Proceedings 10},
  pages={412--426},
  year={2021},
  organization={Springer}
}

@article{akbari2021vatt,
  title={Vatt: Transformers for multimodal self-supervised learning from raw video, audio and text},
  author={Akbari, Hassan and Yuan, Liangzhe and Qian, Rui and Chuang, Wei-Hong and Chang, Shih-Fu and Cui, Yin and Gong, Boqing},
  journal={Advances in neural information processing systems},
  volume={34},
  pages={24206--24221},
  year={2021}
}

@article{summerville2018pcgml,
  author={Summerville, Adam and Snodgrass, Sam and Guzdial, Matthew and Holmgård, Christoffer and Hoover, Amy K. and Isaksen, Aaron and Nealen, Andy and Togelius, Julian},
  journal={IEEE Transactions on Games}, 
  title={Procedural Content Generation via Machine Learning (PCGML)}, 
  year={2018},
  volume={10},
  number={3},
  pages={257-270},
  doi={10.1109/TG.2018.2846639}}

@inproceedings{lucas2019tile,
  title={Tile pattern KL-divergence for analysing and evolving game levels},
  author={Lucas, Simon M and Volz, Vanessa},
  booktitle={Proceedings of the Genetic and Evolutionary Computation Conference},
  pages={170--178},
  year={2019}
}

@inproceedings{guzhov2022audioclip,
  title={Audioclip: Extending clip to image, text and audio},
  author={Guzhov, Andrey and Raue, Federico and Hees, J{\"o}rn and Dengel, Andreas},
  booktitle={ICASSP 2022-2022 IEEE International Conference on Acoustics, Speech and Signal Processing (ICASSP)},
  pages={976--980},
  year={2022},
  organization={IEEE}
}

@article{wang2023connecting,
  title={Connecting multi-modal contrastive representations},
  author={Wang, Zehan and Zhao, Yang and Huang, Haifeng and Liu, Jiageng and Yin, Aoxiong and Tang, Li and Li, Linjun and Wang, Yongqi and Zhang, Ziang and Zhao, Zhou},
  journal={Advances in Neural Information Processing Systems},
  volume={36},
  pages={22099--22114},
  year={2023}
}

@inproceedings{girdhar2023imagebind,
  title={Imagebind: One embedding space to bind them all},
  author={Girdhar, Rohit and El-Nouby, Alaaeldin and Liu, Zhuang and Singh, Mannat and Alwala, Kalyan Vasudev and Joulin, Armand and Misra, Ishan},
  booktitle={Proceedings of the IEEE/CVF conference on computer vision and pattern recognition},
  pages={15180--15190},
  year={2023}
}

@inproceedings{rocamonde2024vision,
  title={Vision-language models are zero-shot reward models for reinforcement learning},
  author={Rocamonde, Juan and Montesinos, Victoriano and Nava, Elvis and Perez, Ethan and Lindner, David},
  booktitle={International Conference on Learning Representations},
  volume={2024},
  pages={28446--28463},
  year={2024}
}

@article{du2023vision,
  title={Vision-language models as success detectors},
  author={Du, Yuqing and Konyushkova, Ksenia and Denil, Misha and Raju, Akhil and Landon, Jessica and Hill, Felix and De Freitas, Nando and Cabi, Serkan},
  journal={arXiv preprint arXiv:2303.07280},
  year={2023}
}

@article{majumdar2022zson,
  title={Zson: Zero-shot object-goal navigation using multimodal goal embeddings},
  author={Majumdar, Arjun and Aggarwal, Gunjan and Devnani, Bhavika and Hoffman, Judy and Batra, Dhruv},
  journal={Advances in Neural Information Processing Systems},
  volume={35},
  pages={32340--32352},
  year={2022}
}

@article{nair2022r3m,
  title={R3m: A universal visual representation for robot manipulation},
  author={Nair, Suraj and Rajeswaran, Aravind and Kumar, Vikash and Finn, Chelsea and Gupta, Abhinav},
  journal={arXiv preprint arXiv:2203.12601},
  year={2022}
}

@article{maaten2008visualizing,
  title={Visualizing data using t-SNE},
  author={Maaten, Laurens van der and Hinton, Geoffrey},
  journal={Journal of machine learning research},
  volume={9},
  number={Nov},
  pages={2579--2605},
  year={2008}
}

@InProceedings{fu2024,
  title = {FuRL: Visual-Language Models as Fuzzy Rewards for Reinforcement Learning},
  author = {Yuwei Fu and Haichao Zhang and Di Wu and Wei Xu and Benoit Boulet},
  booktitle = {Proceedings of the 41st International Conference on Machine Learning},
  year = {2024}
}

@inproceedings{rupp2024g,
  title={G-pcgrl: Procedural graph data generation via reinforcement learning},
  author={Rupp, Florian and Eckert, Kai},
  booktitle={2024 IEEE Conference on Games (CoG)},
  pages={1--8},
  year={2024},
  organization={IEEE}
}

@article{cronbach1951coefficient,
  title={Coefficient alpha and the internal structure of tests},
  author={Cronbach, Lee J},
  journal={psychometrika},
  volume={16},
  number={3},
  pages={297--334},
  year={1951},
  publisher={Springer-Verlag}
}

@article{ruxton2006unequal,
  title={The unequal variance t-test is an underused alternative to Student's t-test and the Mann--Whitney U test},
  author={Ruxton, Graeme D},
  journal={Behavioral Ecology},
  volume={17},
  number={4},
  pages={688--690},
  year={2006},
  publisher={Oxford University Press}
}

@inproceedings{wang2020understanding,
  title={Understanding contrastive representation learning through alignment and uniformity on the hypersphere},
  author={Wang, Tongzhou and Isola, Phillip},
  booktitle={International conference on machine learning},
  pages={9929--9939},
  year={2020},
  organization={PMLR}
}

@article{zhu2024evaluate,
  title={How to evaluate semantic communications for images with vitscore metric?},
  author={Zhu, Tingting and Peng, Bo and Liang, Jifan and Han, Tingchen and Wan, Hai and Fu, Jingqiao and Chen, Junjie},
  journal={IEEE Transactions on Cognitive Communications and Networking},
  volume={10},
  number={5},
  pages={1744--1758},
  year={2024},
  publisher={IEEE}
}

@article{kim2025multi,
  title={Multi-Objective Instruction-Aware Representation Learning in Procedural Content Generation RL},
  author={Kim, Sung-Hyun and Hwang, Geum-Hwan and Baek, In-Chang and Lee, Seo-Young and Kim, Kyung-Joong},
  journal={arXiv preprint arXiv:2508.09193},
  year={2025}
}

\end{document}